%% file: main.tex
\documentclass[runningheads]{llncs}

\usepackage{eccv}

\usepackage{eccvabbrv}

\usepackage{graphicx}
\usepackage{booktabs}

\usepackage[accsupp]{axessibility}  

\usepackage[table]{xcolor}
\usepackage{multirow}
\usepackage{bm}
\usepackage{kotex}
\usepackage{cuted}
\usepackage{wrapfig}
\usepackage{makecell}
\usepackage{tabularx}
\usepackage{amssymb}
\usepackage{amsmath}
\usepackage{dsfont}
\usepackage{array}
\usepackage{pifont}
\usepackage{algorithm}
\usepackage{algpseudocode}
\usepackage{enumitem}
\usepackage{tcolorbox}
\usepackage{pdflscape}
\usepackage{longtable}

\usepackage{hyperref}

\usepackage{orcidlink}

\begin{document}

\title{Articulated Object Reconstruction \\ from Rest-State Observation} 


\author{Daeun Lee \orcidlink{0009-0005-1839-1913},
Jaeah Lee\textsuperscript{*} \orcidlink{0009-0004-2648-8523},
Woosung Kim\textsuperscript{*} \orcidlink{0009-0008-6373-7591},
Haebeom Jung \orcidlink{0009-0005-0602-3005}, \\
and Jaesik Park \orcidlink{0000-0001-5541-409X}
}

\authorrunning{D.~Lee et al.}

\institute{Seoul National University, Republic of Korea}

\maketitle
\def\thefootnote{$^*$}\footnotetext{Indicates equal contribution}
\def\thefootnote{\arabic{footnote}}

\input{preamble.tex}
\input{sec/0_abstract.tex}
\input{sec/1_intro.tex}
\input{sec/2_related_work.tex}

\input{sec/3_method.tex}
\input{sec/4_experiment.tex}
\input{sec/5_conclusion.tex}

\section*{Acknowledgements}
This work was supported by IITP grants (RS-2025-25442338: AI star Fellowship Support Program at SNU~(50\%); RS-2025-02303703: Realworld multi-space fusion and 6DoF free-viewpoint immersive visualization for extended reality~(45\%); RS-2021-II211343: AI Graduate School Program at SNU~(5\%)) funded by the Korea government (MSIT).
We thank Joonghyuk Shin, Minkyun Seo, Jinmo Kim, Youngjoong Kim, Minji Seo, Seunguk Do, and Eun Sun Lee for helpful advice and discussions during the paper.

\clearpage
%
%
\bibliographystyle{splncs04}
\bibliography{main}

\input{sec/X_suppl.tex}

\end{document}

%% file: preamble.tex

\newcommand{\nickname}{\texttt{\textbf{Rest2Art}}\xspace}

\newcommand{\red}[1]{{\color{red}#1}}
\newcommand{\todo}[1]{{\color{red}#1}}
\newcommand{\TODO}[1]{\textbf{\color{red}[TODO: #1]}}

\newcommand{\best}{\cellcolor{tablered}}
\newcommand{\sbest}{\cellcolor{tableorange}}
\newcommand{\tbest}{\cellcolor{yellow}}

\newcommand{\cmark}{\ding{51}}
\newcommand{\xmark}{\ding{55}}

\definecolor{yellow}{rgb}{1, 1, 0.7}
\definecolor{tableorange}{rgb}{1, 0.85, 0.7}
\definecolor{tablered}{rgb}{1, 0.7, 0.7}

\definecolor{oursbg}{RGB}{230,240,250}

\newcommand{\Ra}{\textbf{\color{Dandelion}{Cnx7}}}
\newcommand{\Rb}{\textbf{\color{RubineRed}{8MFH}}}
\newcommand{\Rc}{\textbf{\color{LimeGreen}{bbdj}}}

\newcommand{\annot}[1]{\hfill\textcolor{gray}{\% #1}}









%% file: sec/0_abstract.tex
\begin{figure}
    \centering
    \includegraphics[trim={-2cm 0mm -2cm 0mm}, clip, width=\linewidth]{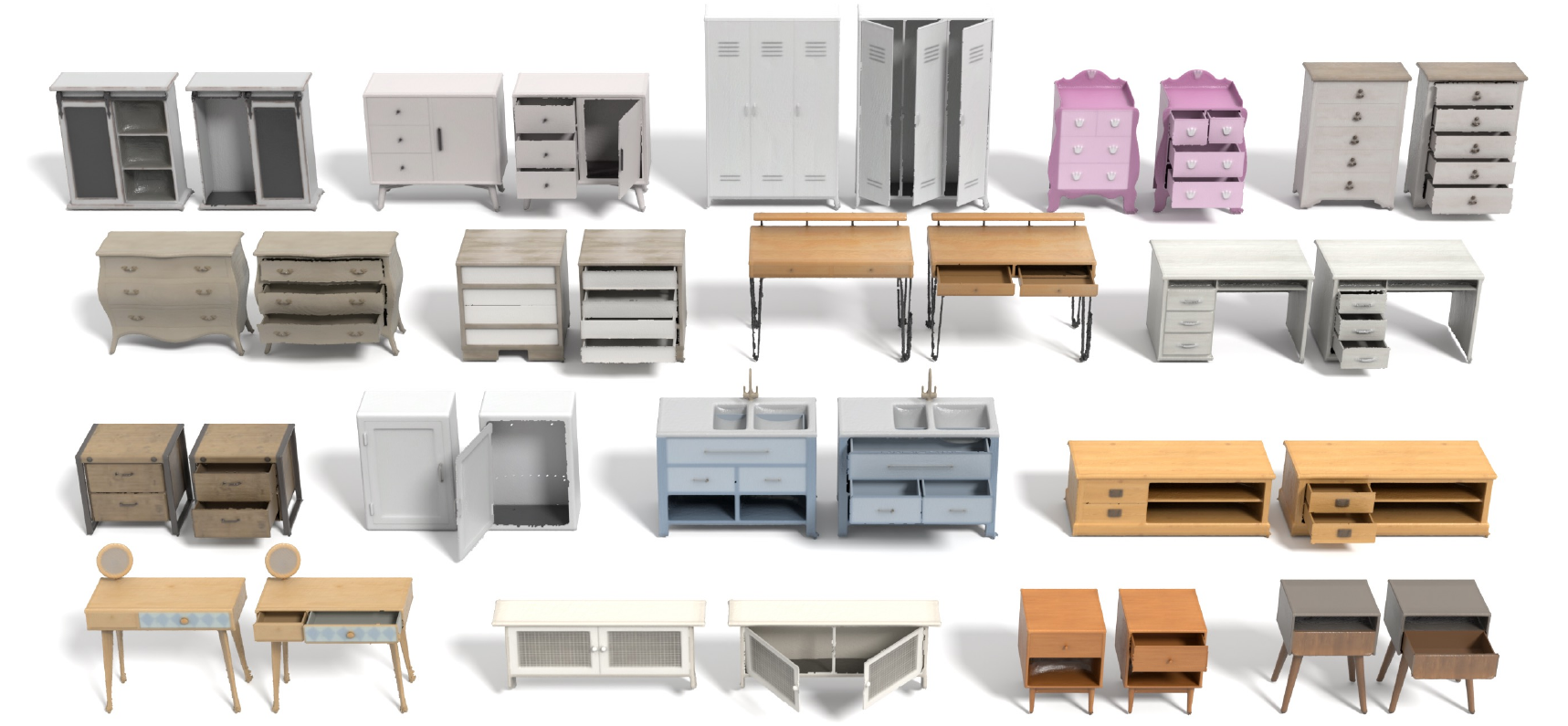}
    \captionof{figure}{
    We propose \nickname, an approach for reconstructing articulated objects with part-level geometry and joint parameters from rest-state observations alone.
    }
    \label{fig:teaser}
    \vspace{-10mm}
\end{figure}

\begin{abstract}
Building interactive digital twins requires recovering both 3D geometry and the kinematic structures that govern how objects articulate.
Yet existing methods for articulated object reconstruction require explicitly observable motion from multiple articulation states.
We introduce a \textit{rest-state} formulation that reconstructs articulated objects from a single closed configuration, an inherently ill-posed setting where geometry, semantics, and motion priors compensate for the absence of motion cues.
Our framework adopts an explicit mesh as an intermediate representation for cross-model verification and fusion, reconciling noisy outputs from vision-language and segmentation models into spatially consistent part structures.
To estimate joint parameters without observed motion, we use a video diffusion model to synthesize articulation hypotheses and validate them through geometric consistency.
Our approach achieves accurate part decomposition and physically plausible articulation, performing competitively with motion-observing reconstruction-based, generation-based, and modular pretrained-model baselines. 
\keywords{Articulated Object
\and Mesh Segmentation
\and Digital Twin}
\end{abstract}

%% file: sec/1_intro.tex
\section{Introduction}
\label{sec:intro}

Humans exhibit a remarkable ability to infer how unfamiliar objects articulate from their appearance alone.
Upon encountering a closed cabinet, we can anticipate how its doors swing or drawers slide before any interaction occurs.
Yet this inference is not always reliable: we sometimes push a door that should slide or misjudge how far a drawer extends.
But once we observe an object in an open state, or watch it being opened, such ambiguity largely disappears.

Existing work on articulated object reconstruction focuses precisely on this well-constrained scenario.
\Cref{tab:intro_articulated_object_reconstruction} summarizes methods across input conditions and intermediate representations.
Some methods~\cite{jiang2022ditto,jiayi2023paris,weng2024dta,liu2025artgs,wu2026reartgs,wu2026reartgspp} require observing objects in carefully selected articulation states and assume prior knowledge of the number of movable parts.
Others relax some of these constraints through feed-forward prediction~\cite{yuan2025larm} or by employing foundation models to estimate joint parameters from a single articulated-state observation, where part displacements remain visible~\cite{mandi2025real2code}.
In these cases, however, the input must contain explicitly visible motion.

\input{tables/1_baselines.tex}

In our work, we aim to replicate ``what humans routinely do'' by inferring how objects articulate from their closed configurations alone.
We call this the \textit{rest-state} setting and focus on openable objects such as furniture, which exhibit diverse multi-part articulation patterns.
In practice, objects are most often encountered in their closed configuration.
It is also the state that can be reliably captured at scale, as it is consistently present in online product photos, internet imagery, and scene-level datasets such as ScanNet~\cite{dai2017scannet, yeshwanthliu2023scannetpp} and Replica~\cite{replica19arxiv}.
Our key observation is that pretrained models can supply rich semantic and motion priors for this task.
A video diffusion model, for instance, can imagine how a closed drawer might open.
However, individual model outputs are often noisy, and each may disagree on which parts move and how they articulate.

\begin{figure*}[t]
    \centering
    \includegraphics[width=0.94\linewidth]{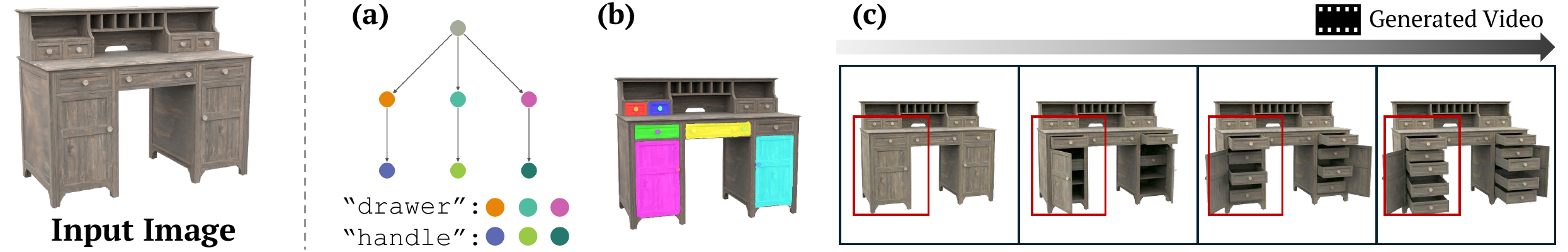}
    \caption{\textbf{Na\"ively applying pretrained models to rest-state articulated objects.} Given a closed cabinet with 2 doors and 7 drawers: (a) a VLM~\cite{openai2025gpt5} predicts an incomplete part hierarchy; (b) segmentation masks~\cite{carion2025sam3segmentconcepts} miss parts in certain views; (c) a video diffusion model~\cite{wan2025} hallucinates nonexistent drawers when opening a door.
    }
    \label{fig:intro_pretrained_model}
\end{figure*}

A Vision-Language Model~(VLM) may predict an incomplete part hierarchy, a segmentation model may miss parts in certain views, and a video diffusion model may hallucinate nonexistent components (\cref{fig:intro_pretrained_model}).
We address this by grounding all predictions in an explicit \textit{mesh}, whose surface connectivity enables spatially consistent part boundaries, reliable occlusion reasoning, and confidence-weighted aggregation of noisy, view-dependent predictions into a coherent 3D result.
This design also makes the framework input-agnostic, as any source that produces a mesh can serve as input, including multi-view reconstruction~\cite{huang20242DGS,kerbl3Dgaussians,svraster}, single-image 3D generation~\cite{xiang2025structured,sam3dteam2025sam3d3dfyimages}, and existing 3D assets.

Based on this insight, we present Rest-to-Articulation~(\nickname), a framework that reconstructs multi-part articulated objects from rest-state observations alone.
We extract a surface mesh from diverse inputs and iteratively co-refine part hierarchies and segmentations using a VLM~\cite{openai2025gpt5,bai2025qwen3} and SAM3~\cite{carion2025sam3segmentconcepts}, using their disagreements as a signal to correct each other.
Validated masks are lifted onto the mesh through confidence-weighted evidence accumulation and refined via graph-based label propagation.
Since no motion is observed, we synthesize plausible articulation sequences using a video diffusion model~\cite{wan2025} to generate motion hypotheses, extract 2D trajectories, and fit rigid-body joint models by minimizing reprojection errors.
The mesh geometry further constrains the joint estimation. 
We penalize axis configurations that cause inter-part penetration and refine the joint alignment against the mesh.
Segmented patches are then completed into volumetric meshes suitable for physical simulation.

To summarize, our contributions are as follows:
\begin{itemize}[label=$\circ$,noitemsep,topsep=0pt,leftmargin=*]
    \item We formulate rest-state articulated object reconstruction as an ill-posed recovery problem and present a complete framework that operates from a single closed configuration without any observed motion.
    \item We design an iterative co-refinement loop between a Vision-Language Model and a segmentation model, where their disagreements drive mutual correction, and lift the validated evidence onto the mesh through confidence-weighted per-part aggregation.
    \item We propose using video diffusion models as a source of motion hypotheses and constrain the resulting joint parameters against the mesh geometry.
    \item We validate across diverse inputs including multi-view captures, single-image 3D generation, and existing 3D assets, demonstrating competitive joint estimation accuracy against both reconstruction and generation baselines.
\end{itemize}

%% file: tables/1_baselines.tex
\begin{table}[t]
\centering
\small
\setlength{\tabcolsep}{4pt} 
\caption{\textbf{Comparison of articulated object reconstruction methods.}
If \textit{\# Parts} is given as a number, it indicates that methods require prior knowledge of the part count up to that value, while \textit{any} denotes that methods discover parts without such priors.}
\resizebox{0.95\linewidth}{!}{%
\begin{tabular}{@{}l
  >{\centering\arraybackslash}p{1.7cm}
  >{\centering\arraybackslash}p{2cm}
  >{\centering\arraybackslash}p{2.1cm}
  >{\centering\arraybackslash}p{1.5cm}
  >{\centering\arraybackslash}p{2.5cm}@{}}
\toprule
\multirow{2}{*}[-2pt]{\textbf{Framework}} & \multirow{2}{*}[-2pt]{\textbf{Task}} & \multicolumn{3}{c}{\textbf{Input}}
& \multirow{2}{*}[-2pt]{\makecell{\textbf{Intermediate}\\\textbf{Representation}}} \\
\cmidrule(lr){3-5}
& & Type & \#~States & \#~Parts & \\
\midrule
Ditto~\cite{jiang2022ditto} & Recon. & PC & 2 & 2 & Neural Field \\
PARIS~\cite{jiayi2023paris} & Recon. & RGBs & 2 & 2 & Neural Field \\
DTA~\cite{weng2024dta} & Recon. & RGB\text{-}Ds & 2 & 3 & Neural Field \\
ArtGS~\cite{liu2025artgs} & Recon. & RGB(-D)s & 2 & any & 3DGS \\
Real2Code~\cite{mandi2025real2code} & Recon. & RGB(-D)s & 1~(articulated) & any & Point Clouds \\
LARM~\cite{yuan2025larm} & Recon. & RGBs & 2 & 2 & Latent \\
REArtGS~\cite{wu2026reartgs} & Recon. & RGBs & 2 & 2 & 3DGS \\
REArtGS++~\cite{wu2026reartgspp} & Recon. & RGBs & 2 & any & 3DGS \\
Articulation in Motion~\cite{ai2026aim} & Recon. & RGBs${+}$video & video & any & 3DGS \\
Articulate AnyMesh~\cite{qiu2025artianymesh} & Recon. & Mesh & 1~(rest) & any & Mesh \\
\midrule
URDFormer~\cite{chen2024urdformer} & Gen. & RGB & 1~(rest) & any & - \\
Singapo~\cite{liu2025singapo} & Gen. & RGB & 1~(rest) & any & - \\
\midrule
\rowcolor{oursbg}
\textbf{Ours} & Recon. & RGBs & 1~(rest) & any & Mesh \\
\bottomrule
\end{tabular}
}
\label{tab:intro_articulated_object_reconstruction}
\end{table}

%% file: sec/2_related_work.tex
\section{Related Work}

\subsubsection{Articulated Object Recovery.}
Recovering articulated objects requires jointly estimating 3D geometry and kinematic structure across movable parts.
Early approaches relied on category-specific priors or predefined kinematic templates~\cite{li2019articulated-pose, heppert2023carto, xue2021omad}, limiting their ability to estimate articulations of unseen objects.
More recent methods adopt per-scene optimization~\cite{jiayi2023paris, zhang2025iaao, liu2025artgs, guo2025articulatedgs, weng2024dta, wu2026reartgs, wu2026reartgspp} or incorporate foundation models for part-level understanding~\cite{le2025articulate, zhang2025iaao}.
Others leverage implicit object-centric representations~\cite{heppert2023carto} and interaction-based cues such as drag points~\cite{li2024dragapart} or interaction videos~\cite{ai2026aim}.
Despite this progress, most methods require multi-state observations, part annotations, or partially open configurations~\cite{mandi2025real2code}, making rest-state reconstruction particularly challenging.
For single rest-state inputs, URDFormer~\cite{chen2024urdformer} and Singapo~\cite{liu2025singapo} leverage 3D generative models to predict articulated geometry from a single image.
Articulate AnyMesh~\cite{qiu2025artianymesh} instead brings generative priors into a reconstruction framework, sequentially chaining pretrained models for open-vocabulary articulation modeling.
In contrast, our approach recovers both accurate geometry and joint parameters from a single closed configuration, without prior knowledge of the number of parts, by iteratively cross-validating foundation model outputs against the reconstructed mesh rather than chaining them sequentially, and constraining joint estimation through mesh-based geometric verification.

\subsubsection{Part Segmentation for Articulated Objects.}
Articulating a rest-state object requires first identifying its movable parts, which is challenging in closed configurations where uniform colors, box-like geometry, and flush boundaries leave little discriminative signal.
Recent zero-shot 3D part segmentation methods lift 2D detections onto point clouds via multi-view voting~\cite{michele2021generative, liu2023partslip, zhou2023partslippp, tang2024samesh}, train feedforward point embeddings on web-scale assets~\cite{ma2025find3d}, or learn continuous 3D feature fields via contrastive distillation~\cite{liu2025partfield}.
In the articulated object domain, Real2Code~\cite{mandi2025real2code} fine-tunes SAM~\cite{kirillov2023sam} with point prompts, but assumes an articulated-state input where displaced parts are visually distinguishable.
Our method addresses this by first co-refining part hierarchies and segmentation masks through iterative cross-model verification, then fusing the validated outputs onto the mesh surface through confidence-weighted multi-view accumulation and graph-based label propagation, without part annotations or training.

\begin{figure*}[t]
    \centering
    \includegraphics[width=\linewidth]{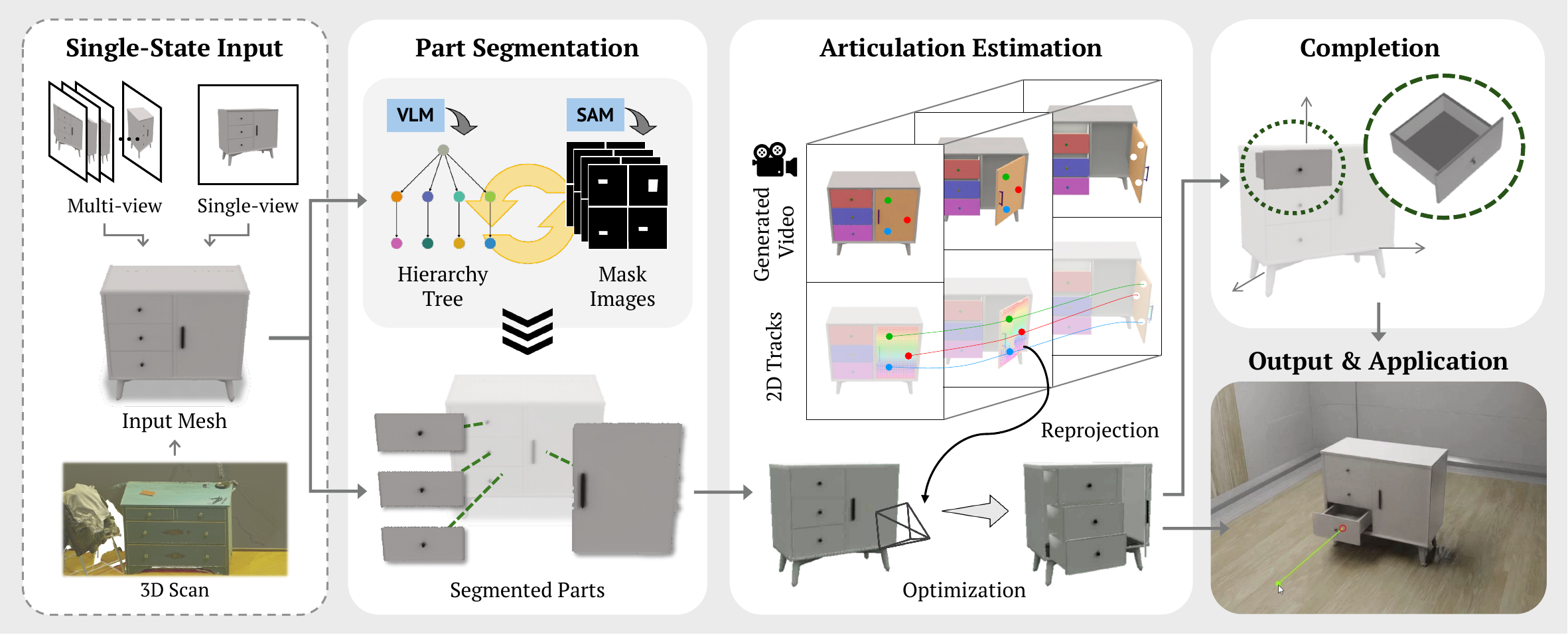}
    \caption{\textbf{Method overview.}
    \nickname~reconstructs articulated objects from rest-state inputs by co-refining part hierarchies and masks, lifting evidence onto the mesh, synthesizing videos for joint estimation, and completing interior geometry.
    }
    \label{fig:overview}
\end{figure*}

%% file: sec/3_method.tex
\section{Method}

Given a rest-state mesh of an articulated object~(\cref{sec:method_problem}), our pipeline proceeds in three stages.
We first identify movable parts through co-refinement of hierarchy predictions and segmentation masks~(\cref{sec:method_part_seg}).
We then estimate joint parameters by synthesizing articulation videos and fitting rigid-body models to extracted trajectories~(\cref{sec:method_art_estim}).
Finally, we complete each part into a volumetric mesh for physics simulation~(\cref{sec:method_part_comp}).
An overview is shown in \cref{fig:overview}.

\subsection{Problem Statement}
\label{sec:method_problem}
We focus on multi-part openable objects, such as furniture, whose closed configuration can be clearly defined.
Formally, given a rest-state object mesh, we output part-wise meshes $\{\mathcal{M}_k\}_{k=1}^{K}$ and joint parameters $\{\Psi_k\}_{k=1}^{K-1}$, where $K$ is the number of parts, automatically determined without prior annotation.

We model the articulated object as a collection of rigid body parts connected by joints, where each part's motion can be represented by an $\text{SE}(3)$ transformation. 
We consider \emph{prismatic} and \emph{revolute} joints, which cover most furniture articulation mechanisms.
For a prismatic joint, we define $\Psi^p = \{\mathbf{v} \in \mathbb{R}^3,\, d \in \mathbb{R}\}$, where $\mathbf{v}$ represents the translation direction and $d$ is the scalar displacement along that direction.
For a revolute joint, we define $\Psi^r = \{\boldsymbol{\omega} \in \mathbb{R}^3,\, \mathbf{q} \in \mathbb{R}^3,\, \theta \in \mathbb{R}\}$, where $\boldsymbol{\omega}$ denotes the direction of the rotation axis, $\mathbf{q}$ is a point on the axis, and $\theta$ indicates the rotation angle.

\subsection{Part Segmentation}
\label{sec:method_part_seg}
Without motion cues, articulation can only be inferred from static geometry and semantics, making part segmentation the central challenge of rest-state recovery.
We decouple segmentation from reconstruction by adopting a \textit{mesh} as the intermediate 3D representation.
Unlike point- or volume-based primitives, the mesh explicitly encodes surface connectivity, which enables precise occlusion reasoning and spatially consistent part boundaries, and serves as the anchor for verifying and fusing outputs from multiple pretrained models throughout this stage.

We obtain an initial mesh $\mathcal{M} = (\mathcal{V}, \mathcal{F})$, where $\mathcal{V}$ and $\mathcal{F}$ denote vertices and faces, respectively.
Our pipeline is reconstruction-agnostic and accepts any mesh-based output, including multi-view reconstruction~\cite{huang20242DGS,kerbl3Dgaussians,svraster}, single-image 3D generation~\cite{sam3dteam2025sam3d3dfyimages,xiang2025structured}, or scene scans.
The mesh is canonically aligned and re-rendered from $N'$ viewpoints chosen to maximize visibility of articulated components.
The view selection strategy is detailed in \cref{sec:supp_view}.

\subsubsection{Co-Refinement of Hierarchy and Part Masks.}
We leverage a VLM~\cite{openai2025gpt5, bai2025qwen3} to propose a candidate hierarchy tree $T$ and SAM3~\cite{carion2025sam3segmentconcepts} to produce per-view part masks $M_v$.
However, na\"ively combining their outputs often fails due to missing parts, spurious detections, and view-dependent segmentation noise.
We therefore introduce an iterative cross-model verification loop that uses disagreements between the two models as a signal for mutual correction.

\begin{figure}[t]
    \centering
    \includegraphics[width=\linewidth]{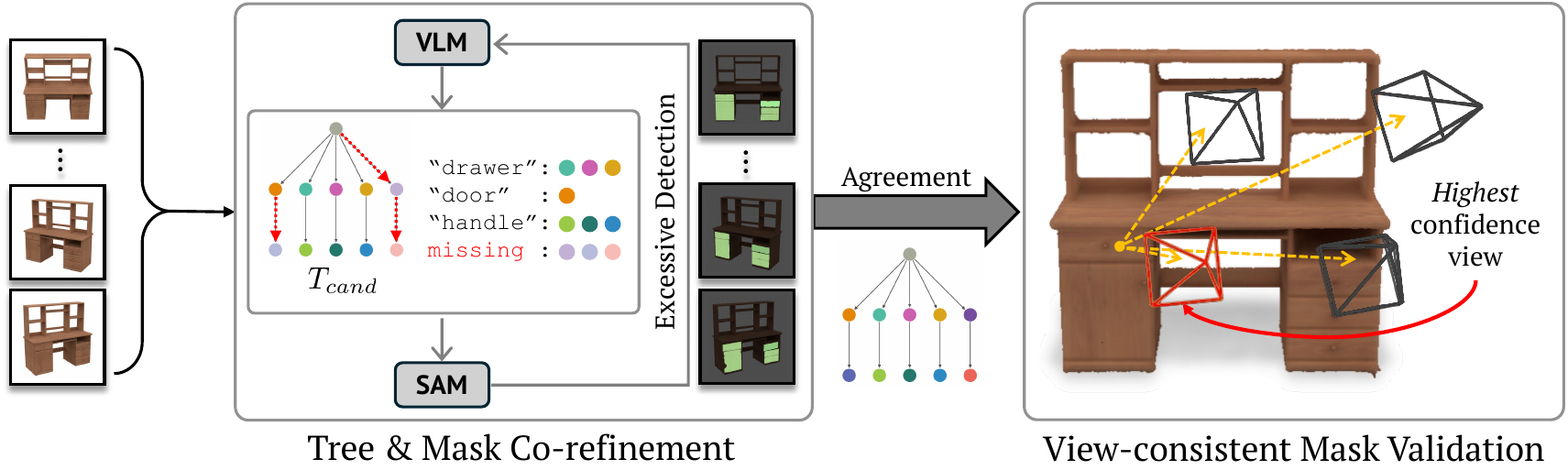}
    \caption{\textbf{Co-refinement of hierarchy and part masks.}
     Disagreements between the VLM hierarchy tree and SAM3 masks drive mutual correction. Validated masks are then back-projected onto the mesh to enforce cross-view consistency.}
    \label{fig:method_corefine}
\end{figure}

From $T$, we extract $C$ semantic prompts $\mathcal{S}(T) = \{s_1, s_2, \ldots, s_C\}$ (\eg, \texttt{drawer}, \texttt{handle}, \texttt{door}), where $C$ is the number of distinct part categories in $T$, along with their expected instance counts $n(s, T)$ for each $s \in \mathcal{S}(T)$.
We query the segmentation model with each prompt $s$ on every view $v$ to obtain a set of masks $M_{v,s}$, and compare the detected count $|M_{v,s}|$ against $n(s, T)$~(\cref{fig:method_corefine}):
\begin{itemize}[label=$\circ$,leftmargin=1.5em]
    \item \textbf{Agreement}
    ($|M_{v,s}| = n(s, T)$ for all $s \in \mathcal{S}(T)$):
    the corresponding masks are cached as validated evidence for this view.

    \item \textbf{Excess detections}
    ($\exists\, s \in \mathcal{S}(T)$ s.t.\ $|M_{v,s}| > n(s, T)$):
    $T$ is incomplete.
    We overlay the detected masks in vivid colors onto the input view, making part boundaries visually distinguishable even on textureless objects, and feed this annotated image back to the VLM to revise $T$.
    Verification then restarts under the updated tree.

    \item \textbf{Missing detections}
    ($\exists\, s \in \mathcal{S}(T)$ s.t.\ $|M_{v,s}| < n(s, T)$):
    rather than pruning $T$, we defer the decision and aggregate evidence across all views before concluding the part is missing.
\end{itemize}
We fix the maximum number of refinement rounds to two, which suffices in practice.
To further enforce cross-view consistency, we exploit the mesh geometry to back-project high-confidence masks into 3D surface points and reproject them onto other views as point prompts for the segmentation model.
A prompted mask is accepted only when it agrees with the cached mask for the same part.

\subsubsection{Confidence-Weighted Evidence Lifting.}
\label{sec:method_evidence_lifting}
Given the verified hierarchy $T$ and per-view masks $\{M_v\}_{v=1}^{N'}$, we lift 2D mask evidence onto the mesh to obtain a globally consistent 3D labeling.
We perform \textit{per-part} view selection and weighting, so each part's labeling is driven by its most reliable observations.

For each part $p$, we rank views by a quality score that jointly considers detection confidence and mask coverage, and select the top-$K$ views $\mathbf{V}_p$.
We rasterize $\mathcal{M}$ from each selected view $v$ using z-buffered rendering~\cite{ravi2020pytorch3d} with back-face culling, producing a per-pixel face index map $\phi_v$.
The \emph{face support set} of mask $M_p^v$ is defined as:
\begin{equation}
    \text{supp}(M_p^v) = \{ f \in \mathcal{F} \mid \exists\, \mathbf{x} \in M_p^v \text{ s.t. } \phi_v(\mathbf{x}) = f \}.
\end{equation}
We accumulate evidence across selected views to obtain a per-face, per-part score:
\begin{equation}
    E(f, p) = \sum_{v \in \mathbf{V}_p} w_p^v \cdot c_p^v \cdot \mathds{1}\big[f \in \text{supp}(M_p^v)\big],
    \label{eq:method_evidence_lifting}
\end{equation}
where $w_p^v$ is the normalized view weight and $c_p^v$ is the detection confidence.
Each face is assigned the part with the maximal accumulated evidence,
\begin{equation}
    \ell(f) = \arg\max_{p} \; E(f, p),
\end{equation}
while faces receiving no support default to the base part.

\subsubsection{Graph-Based Refinement.}
\label{sec:method_graph_refine}
Lifting 2D masks onto the mesh yields a coarse part assignment that still contains three types of errors: boundary inconsistencies from conflicting multi-view evidence, spurious fragments from noisy detections, and under-segmented regions missed by all selected views.
We address these sequentially by exploiting the mesh adjacency graph, which provides the structural connectivity.

We first identify reliable seed regions by computing face-level connected components for each part and retaining only the dominant component per label.
The remaining disconnected fragments, which typically arise from noisy or inconsistent detections, form an unassigned pool $\mathcal{U}$.

To recover under-segmented regions and resolve fragments in $\mathcal{U}$, we propagate labels from seed regions via multi-source shortest-path computation on the vertex adjacency graph.
Each unassigned vertex is assigned the label of the nearest seed part:
\begin{equation}
    \ell(u) = \arg\min_{p} \; d_p(u), \quad u \in \mathcal{U},
\end{equation}
where $d_p(u)$ is the shortest-path distance from $u$ to the nearest seed of part $p$, with ties broken by view quality score.
This propagation respects surface topology, expanding parts along the mesh rather than across geometric discontinuities.

Finally, we verify each propagated label by projecting the relabeled vertex back to its part's best-scoring view, accepting the label only if the projection falls within the corresponding 2D mask.
Vertices that fail this cross-modal check are reassigned to the next nearest part, ensuring that the final labeling is supported by both geometric proximity and 2D evidence.

\subsection{Articulation Estimation}
\label{sec:method_art_estim}
Given the segmented parts and their hierarchy tree $T$, we estimate joint parameters for each articulated part.
In the rest-state setting, no motion is directly observed.
We address this by synthesizing articulation videos through a video diffusion model and recovering joint parameters from the observed 2D trajectories.
Crucially, the synthesized motion serves only as a \emph{source of hypotheses}; the final joint parameters are determined by geometric consistency, not by the fidelity of the generated video.

\subsubsection{Articulation Video Synthesis.}
We overlay the resulting part masks with transparency onto the view with the highest average quality score across parts, explicitly highlighting all movable parts in a single conditioning image for the video model.
We employ Wan2.2~\cite{wan2025} with a LoRA weight from VBVR~\cite{wang2026vbvr}, which encourages spatiotemporal reasoning over physical properties such as continuity and causality.
A VLM~\cite{bai2025qwen3} generates a text prompt describing the expected articulation.
Prompt construction details are provided in \cref{sec:supp_video}.

From the synthesized video, we extract dense 2D point trajectories using CoTracker3~\cite{karaev2025cotracker3}, initializing tracking points on each part's mask region in the first frame.
The mesh provides metric depth at each tracked pixel via rendering, allowing us to unproject the frame-0 positions into 3D camera-frame coordinates as rest-pose anchor points $\{\mathbf{p}_0^j\}_{j=1}^{N_{\mathrm{tr}}}$, where $N_{\mathrm{tr}}$ is the number of tracked points.

Although the generated video may not be geometrically exact, we only require that the \textit{type} and \textit{direction} of motion are approximately correct, since the final joint parameters are fitted against the 3D mesh rather than the video itself.
Extracting point tracks and fitting rigid-body joint models is well established for real motion observations~\cite{werby25arti4d, dharmarajan2026dream2flow}; our setting differs in that no real motion exists, so the video diffusion model must \textit{imagine} plausible articulation while the mesh provides geometric grounding to validate it.

We identify the reliable portion of the synthesized video by truncating tracks at the first frame where the fraction of visible tracks falls below a threshold.
This addresses two issues that both manifest as falling visibility: temporal degradation in generated video, where errors compound over frames and motion becomes implausible~\cite{wan2025}, and accumulated tracker drift.

\subsubsection{Motion Type Classification.}
Before fitting parametric models, we classify each part's motion as revolute, prismatic, or static using only the 2D tracks.
We compute two cues from the truncated track sequence.
First, the coefficient of variation (CV) of per-point displacement magnitudes: revolute motion produces high CV because points at different radii from the rotation axis travel different distances, whereas prismatic motion yields low CV as all points undergo the same translation.
Second, a curvature score that measures the directional change of displacement vectors over time: revolute tracks curve while prismatic tracks remain straight.
Parts with negligible displacement are classified as static.
When both cues agree, the classification is used directly.
In ambiguous cases, we fit both models and select the one with lower reprojection error.

\subsubsection{Joint Parameter Fitting.}
We formulate joint estimation as minimizing the 2D reprojection error between the motion-model prediction and the observed tracks.
For a revolute joint, each rest-pose anchor point $\mathbf{p}_0$ is rotated about an axis $(\hat{\boldsymbol{\omega}}, \mathbf{q})$ by a per-frame angle $\theta_t$:
\begin{equation}
  \mathbf{p}_t
    = R(\hat{\boldsymbol{\omega}},\,\theta_t)\,
      (\mathbf{p}_0 - \mathbf{q})
    + \mathbf{q},
\label{eq:method_revolute}
\end{equation}
where $R(\hat{\boldsymbol{\omega}}, \theta_t)$ denotes the Rodrigues rotation matrix.
For a prismatic joint, the model simplifies to a per-frame scalar displacement $d_t$ along a unit direction $\hat{\mathbf{v}}$:
\begin{equation}
  \mathbf{p}_t = \mathbf{p}_0 + d_t\,\hat{\mathbf{v}}.
\label{eq:method_prismatic}
\end{equation}
In both cases, the predicted 3D positions are projected back to 2D using known camera intrinsics, and the parameters are optimized with a robust loss that down-weights outlier correspondences, followed by a trimmed refit.

The fitted revolute axis, however, inherits ambiguity from monocular depth: multiple 3D rotation axes produce identical 2D projections, making the true rotation direction underdetermined from tracks alone.
We resolve this using the segmented mesh by penalizing axes whose rotation drives the child part into the parent. We further refine $\hat{\boldsymbol{\omega}}$ by softly blending it with the principal direction of the adjacent part boundary, which encodes the physical articulation structure.

To improve robustness against video generation artifacts, we synthesize two articulation videos with different random seeds and fit joint parameters from each independently.
When both fits agree on the motion type, we combine their trajectory evidence and refit jointly, yielding more stable estimates.

\subsection{Mesh Completion}
\label{sec:method_part_comp}
The segmented parts are shell-like patches, but physics simulators and URDF export require volumetric solids.
We convert each part into a closed mesh by offsetting vertices inward along their normals.
Since vertex normals near cut edges are unreliable, we apply boundary-aware normal smoothing to avoid artifacts.

While this solidification suffices for exterior-dominant parts such as doors and knobs, components like drawers require explicit interior geometry, as rest-state inputs limit visibility of interior structures.
For these, we generate a parametric inner tray and blend it with the solidified exterior.
Details of both procedures are provided in \cref{sec:supp_completion}.

%% file: sec/4_experiment.tex
\section{Experiments}
We conduct experiments to validate our rest-state modeling approach from multiple perspectives.
We evaluate part segmentation quality against existing baselines (\cref{sec:exp_seg}), compare joint estimation accuracy with single-state and two-state methods (\cref{sec:exp_joint}), validate our design choices through ablations (\cref{sec:exp_ablation}), and demonstrate generalization to diverse inputs (\cref{sec:exp_diverse_inputs}).

\subsection{Experimental Details}

\subsubsection{Dataset.}
We evaluate on Articulated Containers Dataset~(ACD)~\cite{iliash2026s2o}, a benchmark of articulated furniture objects from Habitat Synthetic Scenes Dataset (HSSD)~\cite{khanna2023hssd} and Amazon Berkeley Objects~(ABO)~\cite{collins2022abo}.
All objects are rendered in closed configuration.
To assess real-world robustness, we additionally evaluate joint estimation on the MultiScan Dataset~\cite{mao2022multiscan}.
We also validate on web images and scanned meshes to evaluate generalization across input modalities.

\subsubsection{Implementation Details.}
For evaluation, we reconstruct meshes from multi-view renderings using 2DGS~\cite{huang20242DGS}, which preserves metric scale consistent with ground truth.
Results with alternative mesh sources are presented in \cref{tab:exp_ablation_recon}.
We use GPT-5.2~\cite{openai2025gpt5} for hierarchy prediction and SAM3~\cite{carion2025sam3segmentconcepts} for mask generation.
For articulation estimation, we generate videos using Wan2.2~\cite{wan2025} with VBVR LoRA~\cite{wang2026vbvr} and extract 2D trajectories using CoTracker3~\cite{karaev2025cotracker3}.

\begin{figure}[t]
    \centering
    \includegraphics[width=\linewidth]{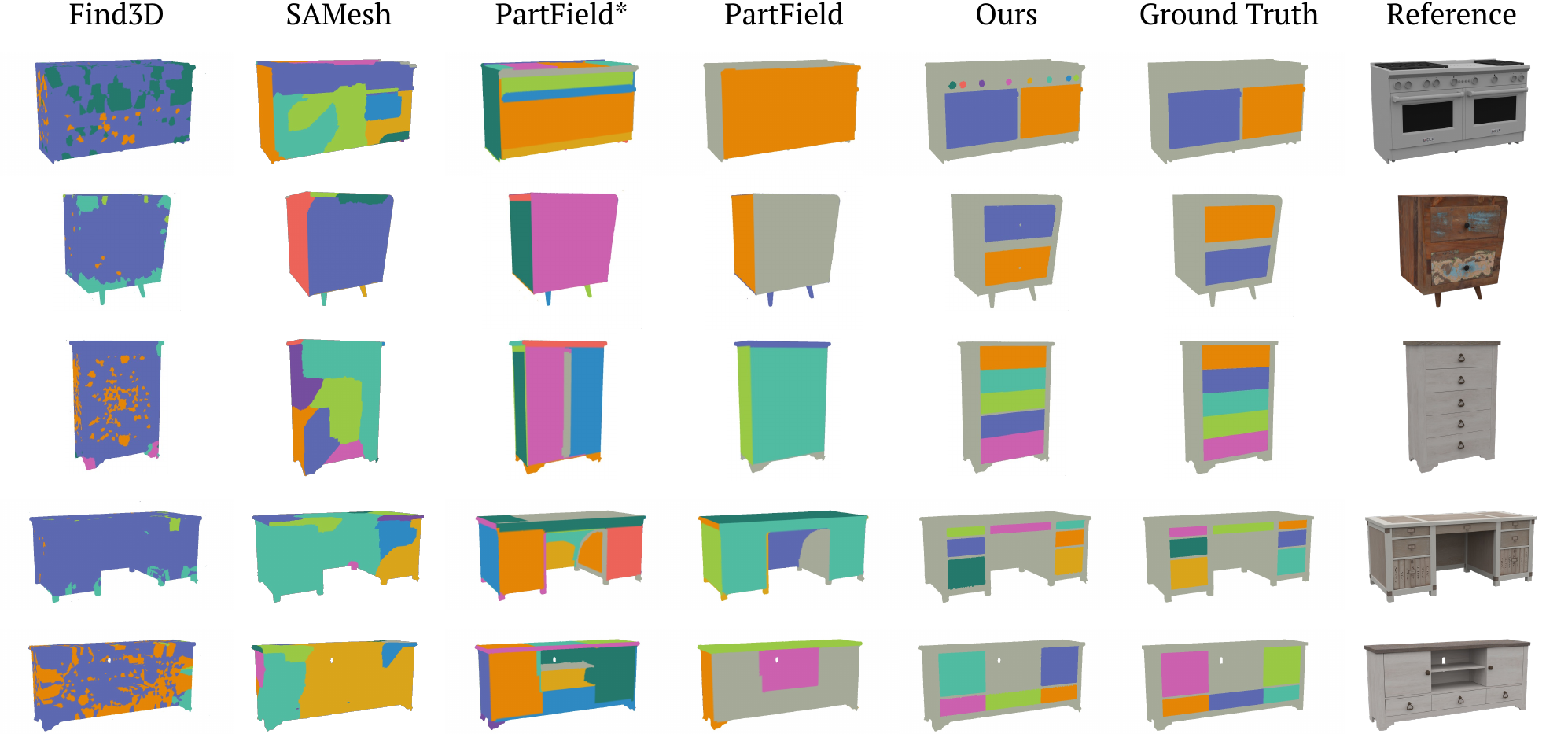}
    \caption{\textbf{Qualitative comparison of part segmentation on ACD dataset~\cite{iliash2026s2o}.} PartField$^{*}$ uses a fixed 20-cluster setting, while PartField uses the ground-truth part count. Our method produces more accurate part boundaries and segments all interactable parts, including knobs~(top row) missing from ground-truth annotations.}
    \label{fig:exp_seg_quali}
\end{figure}

\subsection{Evaluation of Part Segmentation}
\label{sec:exp_seg}

\subsubsection{Baselines.}
We compare with feed-forward 3D segmentation methods PartField~\cite{liu2025partfield}, Find3D~\cite{ma2025find3d}, and the zero-shot mesh segmentation method SAMesh~\cite{tang2024samesh} on the ACD-HSSD dataset~\cite{iliash2026s2o,khanna2023hssd}.
PartField requires the number of parts as input. We evaluate in two settings, one using the ground-truth part count and another with a fixed count of 20 to examine whether over-segmentation can capture functionally meaningful boundaries.
For Find3D, which requires text queries, we use the same prompts as our method for fair comparison. 
Since Find3D operates on point clouds, we sample 5k points from each mesh and transfer predicted labels back to mesh faces following~\cite{liu2025partfield}.
Our method uses the same configuration across all categories and datasets without any training or object-specific tuning.

\subsubsection{Qualitative Results.}
As shown in \cref{fig:exp_seg_quali}, baseline methods struggle on rest-state furniture, where uniform colors, repetitive patterns, and flush boundaries provide minimal discriminative signal for approaches that rely on geometric or visual features alone.
Our method avoids this failure mode by design. 
The co-refinement loop ensures part counts and masks are cross-validated between the VLM and segmentation model before any evidence is committed to the mesh. 
The confidence-weighted lifting then discards unreliable views rather than averaging them.
Quantitative results are provided in \cref{sec:supp_seg}.

\input{tables/4_3_quanti_synthetic.tex}

\subsection{Evaluation of Joint Estimation}
\label{sec:exp_joint}

\subsubsection{Baselines.}
We compare against reconstruction-based baselines.
Among them, only Articulate AnyMesh~\cite{qiu2025artianymesh} shares our rest-state setup, operating directly on our 2DGS-reconstructed mesh.
ArtGS~\cite{liu2025artgs}, REArtGS~\cite{wu2026reartgs}, and REArtGS${++}$~\cite{wu2026reartgspp} are two-state methods that leverage explicit motion observations from paired multi-view images of distinct articulation states.
We synthesize such images by selecting the last physically plausible frame from a generated video, producing multi-view articulated-state images via Qwen-Image-Edit~\cite{wu2025qwenimagetechnicalreport}, estimating camera parameters with MASt3R-SfM~\cite{mast3r_eccv24, duisterhof2025mastrsfm}, and filtering out low-confidence views.
Articulation in Motion~\cite{ai2026aim} additionally requires an interaction video, for which we use a video generated by the same VDM.
Details are provided in \cref{sec:supp_twostate}.

We additionally compare with URDFormer~\cite{chen2024urdformer} and Singapo~\cite{liu2025singapo}, generation-based approaches that can be directly applied in the rest-state setting without requiring motion observations.
Since both methods produce articulated structures at a normalized scale, we estimate a scale factor for each predicted axis to align it with the ground-truth geometry before evaluation.
Following prior work~\cite{jiayi2023paris,weng2024dta,liu2025artgs}, we report axis angular error (Axis Ang.), axis positional error (Axis Pos.), and joint type accuracy (Type Acc.).
Axis angular and positional errors are computed only for parts where the predicted joint type is correct.

\begin{figure*}[t]
    \centering
    \includegraphics[width=\linewidth]{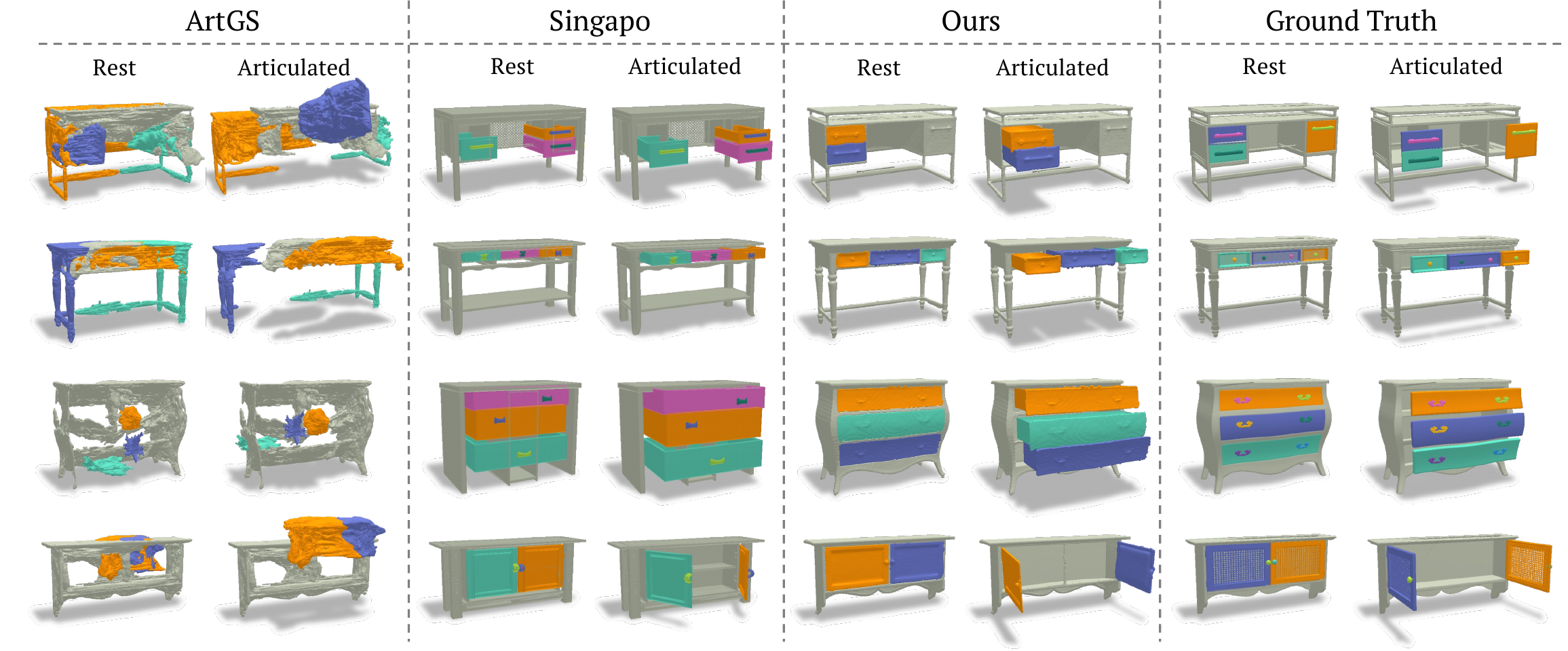}
    \caption{\textbf{Qualitative comparison of joint estimation on ACD dataset~\cite{iliash2026s2o}.}
    Each example shows the predicted rest and articulated states.
    }
    \label{fig:exp_2states_quali}
\end{figure*}

\subsubsection{Results on Synthetic Data.}
We report the quantitative~(\Cref{tab:exp_joint_estimation}) and qualitative~(\cref{fig:exp_2states_quali}) comparison.
Articulate AnyMesh~\cite{qiu2025artianymesh}, which shares our rest-state setup, applies pretrained models sequentially per module and is therefore sensitive to noisy individual predictions, causing it to struggle on objects with subtle articulation boundaries.
ArtGS~\cite{liu2025artgs}, REArtGS~\cite{wu2026reartgs}, and REArtGS${++}$~\cite{wu2026reartgspp} struggle with the synthesized multi-state input, as reconstruction-based two-state methods require precise geometric consistency between states.
In particular, even the static base part is often detected as moving due to slight misalignment between the synthesized states, causing the entire reconstruction to collapse.
Articulation in Motion~\cite{ai2026aim}, designed for real interaction videos, exhibits a similar mismatch with our synthesized video input.
This highlights that rest-state recovery requires robustness to noisy intermediate outputs, whether from pretrained models or motion estimates, which our framework achieves by grounding all evidence in the mesh geometry.

The generative methods benefit from producing structures in the canonical space with axis-aligned joint parameters, yielding low angular error after alignment.
This advantage, however, comes at the cost of metric fidelity, as their outputs require post-hoc scale alignment for meaningful evaluation.

Our method achieves the highest type accuracy among methods directly applicable to rest-state inputs on both datasets, and competitive axis estimation overall, despite recovering joint parameters directly from optimization on generated trajectories without axis-aligned priors.
This demonstrates that the combination of video-based hypothesis sampling and mesh-grounded geometric fitting provides a viable path for rest-state articulation estimation at metric scale.

\subsubsection{Results on Real-World Data.}
\input{tables/4_3_quanti_real.tex}
We further evaluate joint estimation on MultiScan~\cite{mao2022multiscan}, which provides real-world RGB-D scans of indoor scenes with articulated furniture.
We preprocess scene-level annotations into per-instance articulated objects with their RGB views; details are provided in \cref{sec:supp_multiscan}.
Since MultiScan naturally provides two-state captures per object, we use them directly as input to two-state reconstruction baselines~\cite{wu2026reartgs,wu2026reartgspp}. 
For single-state methods, including Articulate AnyMesh~\cite{qiu2025artianymesh} and ours, we use both available states as input and report the result with the lower reprojection error.

Despite noisy real-world inputs, our method recovers relatively accurate joints (\Cref{tab:exp_joint_real}), outperforming reconstruction baselines on type accuracy and angular, and thereby demonstrating the robustness of our mesh-grounded formulation under real-world noise.

\subsection{Ablation Study}
\label{sec:exp_ablation}

\input{tables/4_4_ablation_corefine.tex}

\subsubsection{Co-Refinement Model Selection.}
\Cref{tab:exp_ablation_model} validates our co-refinement loop on the ACD dataset~\cite{iliash2026s2o, collins2022abo, khanna2023hssd}, reporting hierarchy tree accuracy and per-part count accuracy.
A VLM~\cite{openai2025gpt5, bai2025qwen3} alone predicts plausible hierarchies but frequently miscounts parts, as rest-state objects offer minimal visual cues for distinguishing individual components.
A segmentation model~\cite{carion2025sam3segmentconcepts, ren2024grounded, ravi2024sam2segmentimages} alone discovers parts but cannot infer hierarchical relationships without semantic grounding.
Combining both in the iterative verification loop yields the best performance on both metrics, confirming that the two models provide complementary signals that are only fully realized through mutual correction.
Additional co-refinement analysis is provided in \cref{sec:supp_convergence}.

\input{tables/4_4_ablation_recon.tex}

\subsubsection{Mesh Reconstruction Backend.}
\Cref{tab:exp_ablation_recon} compares different reconstruction backends, measuring both segmentation quality and joint estimation accuracy.
For IoU computation between meshes with different topologies, we sample 100k surface points per part and measure overlap based on nearest-neighbor distances.

\begin{figure}[t]
    \centering
    \includegraphics[width=\linewidth]{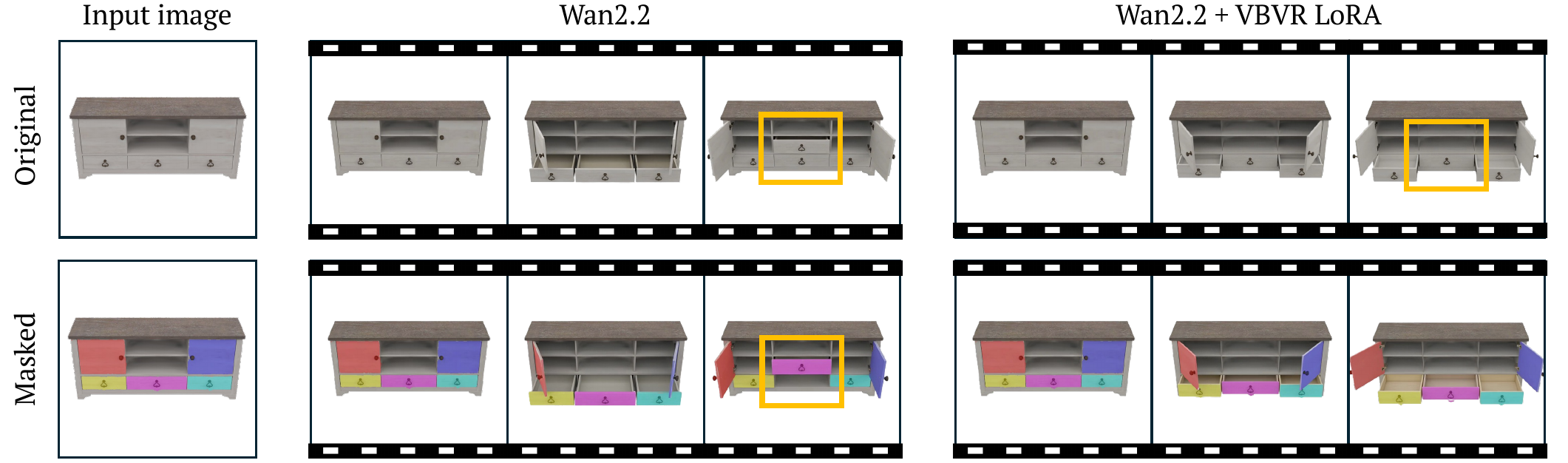}
    \caption{\textbf{Qualitative ablation on video generation.} 
    We compare Wan2.2~\cite{wan2025} with and without VBVR LoRA~\cite{wang2026vbvr}, conditioned on the original image (top) and our mask overlay (bottom). 
    Yellow boxes highlight implausible motion. 
    }
    \label{fig:exp_ablation_video}
\end{figure}

\subsubsection{Video Generation Design.} 
We ablate two design choices in our articulation video synthesis (\cref{fig:exp_ablation_video}).
Wan2.2~\cite{wan2025} alone frequently generates implausible motion such as deforming or hallucinating parts.
VBVR LoRA~\cite{wang2026vbvr} improves plausibility, but mask overlay is necessary to guide the model toward parts with subtle boundaries, reducing hallucination from 23.1\% to 11.5\% and raising the accurate articulation rate from 69.2\% to 80.8\% on ACD-HSSD~\cite{khanna2023hssd}.
Additional analysis of video generation behavior is provided in \cref{sec:supp_vdm}.

\subsection{Generalization to Diverse Inputs} 
\label{sec:exp_diverse_inputs} 

\begin{figure}[t]
    \centering
    \includegraphics[width=\linewidth]{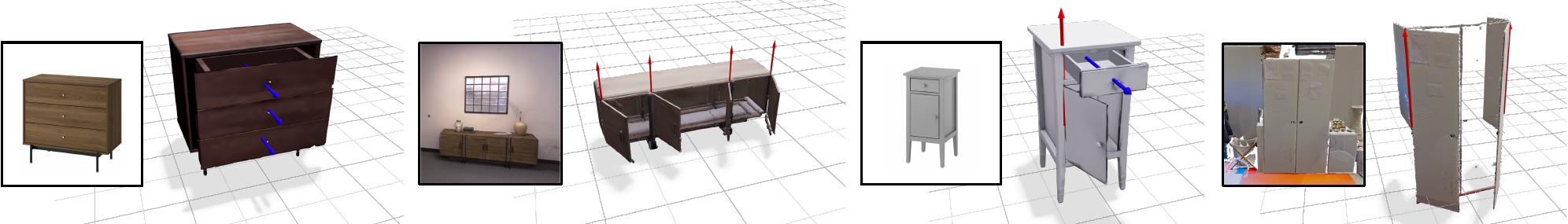}
    \caption{\textbf{Generalization to Diverse Inputs.}
    Our method applies to single-image inputs and scanned meshes, in addition to multi-view captures.
    }
\label{fig:exp_diverse_inputs}
\end{figure}

Since our framework is input-agnostic, it extends beyond multi-view captures.
For single-image inputs, we replace the multi-view reconstruction stage with an image-to-3D generative model~\cite{xiang2025structured,sam3dteam2025sam3d3dfyimages}; for scanned meshes, the pipeline applies directly without modification.
\Cref{fig:exp_diverse_inputs} presents results on online product photos, an indoor scene from Replica~\cite{replica19arxiv}, and scanned meshes from ScanNet${++}$~\cite{yeshwanthliu2023scannetpp}, demonstrating that our formulation generalizes across diverse inputs and real-world imagery.
Additional results are provided in \cref{sec:supp_single_state,sec:supp_realworld}.

%% file: tables/4_3_quanti_synthetic.tex
\begin{table}[t]
    \centering
    \small
    \caption{\textbf{Joint estimation comparison on the ACD dataset~\cite{iliash2026s2o}.}
    Ang., Pos., and Type denote axis angular error~($^\circ$), position error~(dm), and type accuracy~(\%), respectively.
    $\circlearrowright$ denotes our co-refinement strategy, where the co-refined hierarchy tree is used as input.
    $^\dagger$REArtGS requires a single ground-truth joint type per object as input.
    Cell highlights denote ranks: \colorbox{tablered}{best}, \colorbox{tableorange}{second}, \colorbox{yellow}{third}.
    }
    \setlength{\tabcolsep}{4pt}
    \resizebox{\linewidth}{!}{
    \begin{tabular}{lcc|ccc|ccc}
    \toprule
    & \multicolumn{2}{c|}{Setting} & \multicolumn{3}{c|}{ACD-HSSD~\cite{khanna2023hssd}} & \multicolumn{3}{c}{ACD-ABO~\cite{collins2022abo}} \\
    \cmidrule(lr){2-3} \cmidrule(lr){4-6} \cmidrule(lr){7-9}
    Method & Task & State
     & Ang.$\downarrow$ & Pos.$\downarrow$ & Type$\uparrow$
     & Ang.$\downarrow$ & Pos.$\downarrow$ & Type$\uparrow$ \\
    \midrule
    URDFormer~\cite{chen2024urdformer} & Gen. & rest-state & 14.59 & 6.52 & 16.23 & 13.64 & 6.11 & 19.12 \\
    Singapo~\cite{liu2025singapo} & Gen. & rest-state & \best 1.97 & 3.21 & 59.42 & \best 0.13 & 2.63 & 41.47 \\
    Singapo~\cite{liu2025singapo} + $\circlearrowright$ & Gen. & rest-state & \sbest 2.24 & 2.85 & \tbest 61.21 & \sbest 0.14 & 2.33 & \tbest 67.94 \\
    \midrule
    ArtGS~\cite{liu2025artgs} & Recon. & two states & 53.12 & \tbest 1.37 & 57.71 & 46.29 & 0.96 & 60.76 \\
    REArtGS$^\dagger$~\cite{wu2026reartgs} & Recon. & two states & 58.02 & 6.95 & 33.30 & 59.20 & 6.77 & 35.61 \\
    REArtGS++~\cite{wu2026reartgspp} & Recon. & two states & 41.82 & 2.87 & 59.38 & 45.92 & 1.23 & \best 81.28 \\
    Articulation in Motion~\cite{ai2026aim} & Recon. & video & 30.29 & N/A & 25.09 & 19.74 & \tbest 0.81 & 41.90 \\
    Articulate AnyMesh~\cite{qiu2025artianymesh} & Recon. & rest-state & \tbest 11.35 & \sbest 1.31 & \sbest 67.12 & 25.65 & \best 0.08 & 38.94 \\
    \textbf{Ours} & Recon. & rest-state & \tbest 11.35 & \best 0.85 & \best 74.49 & \tbest 4.78 & \sbest 0.20 & \sbest 73.38 \\
    \bottomrule
    \end{tabular}
    }
    \label{tab:exp_joint_estimation}
\end{table}

%% file: tables/4_3_quanti_real.tex
\begin{table}[t]
    \centering
    \small
    \caption{\textbf{Joint estimation on the MultiScan dataset~\cite{mao2022multiscan}.}
    Cov.~(\%) denotes the percentage of evaluable scenes, over which Ang., Pos., and Type are computed.
    Remaining metrics and cell highlights follow \cref{tab:exp_joint_estimation}.}
    \label{tab:exp_joint_real}
    \setlength{\tabcolsep}{4pt}
    \resizebox{0.6\linewidth}{!}{
    \begin{tabular}{l|cccc}
    \toprule
    Method & Cov.\,$\uparrow$ & Ang.\,$\downarrow$ & Pos.\,$\downarrow$ & Type\,$\uparrow$ \\
    \midrule
    REArtGS$^\dagger$~\cite{wu2026reartgs}        & 31.91 & \tbest 51.93 & 29.60 & \sbest 64.40 \\
    REArtGS++~\cite{wu2026reartgspp}              & \tbest 36.17 & 62.72 & 6.85 & \tbest 48.74 \\
    Articulate AnyMesh~\cite{qiu2025artianymesh}  & \best 61.70 & \sbest 30.51 & \tbest 3.45 & 12.63 \\
    \textbf{Ours}                                 & \sbest 48.94 & \best 17.23 & \best 0.26 & \best 65.78 \\
    \bottomrule
    \end{tabular}
    }
\end{table}

%% file: tables/4_4_ablation_corefine.tex
\begin{table}[t]
    \centering
    \caption{\textbf{Ablation on model selection for co-refinement($\circlearrowright$) on ACD dataset.}
        $\mathcal{T}$ Acc. and \#~Parts Acc. denote hierarchy tree and per-part count accuracy($\%$).
        $\dagger$ denotes the use of ground-truth part labels for text grounding.
    }
    \setlength{\tabcolsep}{4pt}
    \resizebox{0.83\linewidth}{!}{
        \begin{tabular}{@{}p{6cm}c|cc@{}}
        \toprule
        Method & Co-Refine & $\mathcal{T}$ Acc.\,$\uparrow$ & \#~Parts Acc.\,$\uparrow$ \\
        \midrule
        GPT-5.2~\cite{openai2025gpt5}           & --  & 62.3\% & 59.4\% \\
        Qwen3-VL~\cite{bai2025qwen3}         & --  & 52.2\% & 47.8\% \\
        SAM3$^\dagger$~\cite{carion2025sam3segmentconcepts}  & --  & -- & 	82.6\% \\
        Grounded-SAM2$^\dagger$~\cite{ren2024grounded, ravi2024sam2segmentimages}  & --  & -- & 43.5\% \\
        \midrule
        GPT-5.2 + SAM3 & \xmark & 62.3 & 59.4\\
        Qwen3-VL + SAM3 & \xmark & 52.2 & 47.8 \\
        Qwen3-VL + Grounded-SAM2 & \xmark & 53.6 & 46.4 \\
        \midrule
        \rowcolor{oursbg}
        GPT-5.2 $\circlearrowright$ SAM3~(Ours)    & \cmark & 72.5\,\tiny(+10.2) & 59.4\,\tiny(+0.0) \\
        \rowcolor{oursbg}
        Qwen3-VL $\circlearrowright$ SAM3~(Ours) & \cmark & 72.5\,\tiny(+20.3) & 63.8\,\tiny(+16.0) \\
        \rowcolor{oursbg}
        Qwen3-VL $\circlearrowright$ Grounded-SAM2~(Ours) & \cmark & 60.9\,\tiny(+7.3) & 47.8\,\tiny(+1.4) \\
        \bottomrule
        \end{tabular}
    }
    \label{tab:exp_ablation_model}  
\end{table}

%% file: tables/4_4_ablation_recon.tex
\begin{table}[t]
    \centering
    \small
    \begin{minipage}[c]{0.48\linewidth}
        \centering
        \caption{\textbf{Ablation on mesh reconstruction methods.}}
        \label{tab:exp_ablation_recon}
        \resizebox{0.98\linewidth}{!}{
        \begin{tabular}{lcc|cc}
        \toprule
        \multirow{2}{*}{} & \multicolumn{2}{c|}{Segmentation} & \multicolumn{2}{c}{Joint Estimation} \\
         & IoU $\uparrow$ & mAP $\uparrow$ & Axis Dir $\downarrow$ & Axis Pos $\downarrow$ \\
        \midrule
        3DGS~\cite{kerbl3Dgaussians}          & 0.30 & 0.83 & 1.69 & 0.154 \\
        SVRaster~\cite{svraster}              & 0.83 & 0.96 & 1.86 & 0.146 \\
        \rowcolor{oursbg}
        \textbf{2DGS~\cite{huang20242DGS}}    & 0.97 & 0.99 & 3.67 & 0.154 \\
        \bottomrule
        \end{tabular}
        }
    \end{minipage}%
    \hfill
    \begin{minipage}[c]{0.48\linewidth}
        \centering
        \includegraphics[trim={0mm -5mm 0mm -10mm}, clip, width=\linewidth]{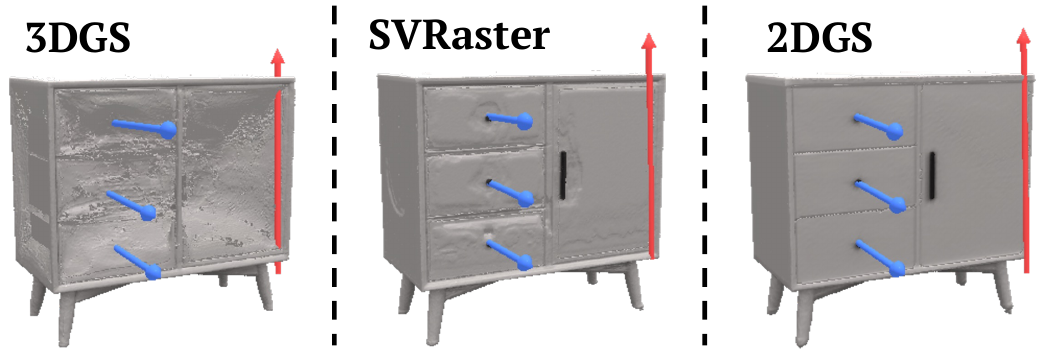}
        \captionof{figure}{\textbf{Qualitative ablation on mesh reconstruction methods.}}
        \label{fig:exp_ablation_recon}
    \end{minipage}
\end{table}

%% file: sec/5_conclusion.tex
\section{Conclusion}
\label{sec:conclu}
We present \nickname, a framework that reconstructs articulated objects from rest-state observations alone by cross-validating pretrained model outputs against explicit mesh geometry.
This demonstrates that models with inconsistent outputs can yield robust results through structured geometric verification.
Our reconstruction-based formulation preserves metric scale and geometric fidelity, producing digital replicas with physically meaningful joint parameters for applications such as interactive simulation and spatial planning.
Extending to more complex kinematic structures and improving the physical plausibility of video generation models are promising future directions.

%% file: sec/X_suppl.tex
\setcounter{page}{1}

\renewcommand{\thesection}{\Alph{section}}
\renewcommand{\thetable}{\Alph{table}}
\renewcommand{\thefigure}{\Alph{figure}}
\setcounter{section}{0}
\setcounter{table}{0}
\setcounter{figure}{0}

\renewcommand{\theHsection}{supp.\arabic{section}}
\renewcommand{\theHsubsection}{supp.\arabic{section}.\arabic{subsection}}
\renewcommand{\theHsubsubsection}{supp.\arabic{section}.\arabic{subsection}.\arabic{subsubsection}}
\renewcommand{\theHtable}{supp.\arabic{table}}
\renewcommand{\theHfigure}{supp.\arabic{figure}}

\section{Implementation Details}
This section provides additional implementation details for each stage of \nickname.
\begin{itemize}[label=$\circ$,noitemsep,topsep=0pt,leftmargin=*]
    \item \Cref{sec:supp_algo} lists the key implementation parameters.
    \item \Cref{sec:supp_view} describes the view selection and canonical alignment strategy.
    \item \Cref{sec:supp_tree} details the hierarchy tree construction and co-refinement procedure.
    \item \Cref{sec:supp_video} describes the video generation model and prompting strategy.
    \item \Cref{sec:supp_track} details track extraction and visibility-gated filtering.
    \item \Cref{sec:supp_completion} describes the mesh completion and solidification procedure.
    \item \Cref{sec:supp_runtime} reports runtime and memory analysis.
\end{itemize}

\subsection{Key Parameters}
\label{sec:supp_algo}
\Cref{tab:supp_impl_params} lists the key implementation parameters.
\input{tables/x_supp_param.tex}

\subsection{View Selection Strategy}
\label{sec:supp_view}

We first align the mesh's oriented bounding box (OBB) to a Blender-convention coordinate system ($+z$ up, $-y$ frontal), then re-render it from $N'$ viewpoints centered around the frontal direction to maximize visibility of articulated components.
When the mesh is produced by a Novel View Synthesis~(NVS)-based pipeline (\eg, 2DGS~\cite{huang20242DGS}, 3DGS~\cite{kerbl3Dgaussians}, SVRaster~\cite{svraster}), we reuse the same renderer.
For standalone mesh inputs, we use Pytorch3D renderer~\cite{ravi2020pytorch3d} and apply the fixed camera configuration in \cref{tab:supp_impl_params}.

After obtaining per-view masks $\{M_{v,s}\}$, each view $v$ is scored per part $p$ by detection confidence weighted by log mask area, preventing views with disproportionately large masks from dominating.
For parent parts, the score is further boosted by the fraction of child parts visible in that view, encouraging selection of views that expose the full part hierarchy.
The normalized view weight $w^v_p$ in \cref{eq:method_evidence_lifting} is the ratio of each view's score to the best-scoring view for part $p$.

\subsection{Hierarchy Tree Construction}
\label{sec:supp_tree}

We query GPT-5.2~\cite{openai2025gpt5} with all $N'$ re-rendered views and select the prediction with the highest part count as the initial hierarchy tree $T$, following the hierarchy graph prediction approach of Singapo~\cite{liu2025singapo}.
We use temperature $0.4$ and 6-shot in-context examples.
During co-refinement, the same prompt template is reused with the mask overlay image, so the model can visually observe the segmented parts and revise its count accordingly.
Parts absent from the final segmentation are pruned from $T$ at the end of co-refinement.
The system prompt is as follows:

\input{algorithms/supp_vlm_prompt_box.tex}

\subsection{Video Generation Details}
\label{sec:supp_video}

\subsubsection{Articulation Video Synthesis.}
We generate articulation videos using Wan2.2-I2V-A14B~\cite{wan2025} with VBVR LoRA~\cite{wang2026vbvr, kijai2025vbvrlora} in an image-to-video (I2V) setting.
The conditioning image is a composite of the front-view rendering with part masks blended in distinct colors ($\alpha = 0.35$).
We produce 81 frames at $960 \times 960$ resolution, which are subsequently downsampled to match the rendering resolution.
Inference is accelerated via LightX2V~\cite{lightx2v} using a 4-step distilled schedule with guidance scale 3.5, with three LoRA adapters applied at strength 1.0 across high- and low-noise phases.
We synthesize $R = 2$ videos with different random seeds.

\subsubsection{Prompt Generation.}
We use Qwen3-VL~\cite{bai2025qwen3} to generate a text prompt from the overlay image using the following instruction template. 
A fixed suffix \texttt{``No camera movement. Camera locked.''} is appended to suppress camera drift.

\input{algorithms/supp_vdm_prompt.tex}

\subsubsection{Multi-Seed Consensus.}
Each seed is processed independently through tracking and joint fitting.
For each part, the best seed is selected by a score combining median displacement and reprojection error.
If a part is classified as static across all $R$ seeds, two additional videos are generated as a fallback to recover missed articulations.
A detailed breakdown of video generation outcomes and failure modes is provided in \cref{sec:supp_vdm} and \cref{sec:supp_failure}.

\subsection{Track Extraction and Filtering}
\label{sec:supp_track}

Tracking points are initialized on a dense grid within the eroded part mask of the first frame, where mask erosion (2\% of the bounding box short side) removes unreliable boundary points.
CoTracker3~\cite{karaev2025cotracker3} performs bidirectional tracking from frame 0.
Tracks are truncated at the first frame where the visible fraction drops below $\tau_\text{vis} = 0.5$ relative to frame 0, jointly filtering degraded video frames and accumulated tracker drift.

\subsection{Mesh Completion Details}
\label{sec:supp_completion}

The interior geometry of articulated objects is inherently unobservable from rest-state images, making true reconstruction ill-posed.
Rather than inferring hidden geometry, we focus on volumetrizing each part into a watertight solid suitable for physical simulation and robotic manipulation.
We complete each open part surface through two steps: solidification and, for drawer parts, interior tray generation.

\subsubsection{Solidification.}
We offset the part surface inward by a base thickness of $\delta = 0.015$\,m to generate an inner surface, then connect boundary edges with side walls to produce a closed solid.
To handle unreliable normals near cut edges, we apply boundary-aware normal smoothing that progressively blends normals within a few hops of the boundary before offsetting.
To avoid interpenetration between adjacent parts, the per-vertex thickness is adaptively clamped to $\min(\delta,\; 0.45 \cdot d_i)$, where $d_i$ is the distance to the nearest vertex of a neighboring part, with a minimum of 0.5\,mm.
Parts that are already watertight are left unchanged.

\subsubsection{Drawer Interior Generation.}
For drawer parts, we additionally generate an interior tray to produce a geometrically plausible hollow structure.
The tray is extruded inward (opposite to the front face normal) with each dimension scaled to $0.75\times$ the corresponding drawer extent, where depth is measured as the available space $d_\text{avail}$ between the drawer back face and the base part boundary.
Wall thickness is set equal to $\delta$.

\input{tables/x_supp_runtime.tex}

\subsection{Runtime and Memory Analysis}
\label{sec:supp_runtime}

We report per-stage runtimes and GPU memory usage measured on the ACD dataset~\cite{iliash2026s2o} using a single NVIDIA H200 GPU.
\Cref{tab:supp_runtime} summarizes the mean runtime per stage, and \cref{fig:supp_vram} reports the per-stage GPU memory profile. Each entry includes the model loading time.
Mesh reconstruction is performed by an external NVS pipeline~(\eg, 2DGS~\cite{huang20242DGS}) and is excluded from the \nickname runtime.
The dominant bottleneck is video generation, which accounts for approximately half of the total runtime and dominates peak GPU memory.
Mesh completion runtime scales with part count and mesh complexity.
Co-refinement reaches up to 72\,s when VLM re-inference is triggered.
Tracking and joint fitting are lightweight, each completing within 25\,s across all tested scenes.
When static recovery is triggered, two additional videos are generated, adding approximately 110\,s of inference time to the total.

\section{Experimental Details}
This section provides additional experimental details for \nickname.
\begin{itemize}[label=$\circ$,noitemsep,topsep=0pt,leftmargin=*]
    \item \Cref{sec:supp_twostate} details the two-state input construction procedure for reconstruction methods requiring two-state input~\cite{liu2025artgs,wu2026reartgs,wu2026reartgspp,ai2026aim}.
    \item \Cref{sec:supp_multiscan} details the preprocessing procedure for the MultiScan~\cite{mao2022multiscan} real-world evaluation.
\end{itemize}

\subsection{Two-State Input Construction}
\label{sec:supp_twostate}

\subsubsection{Why Direct Comparison Is Infeasible.}
Two-state reconstruction methods~\cite{liu2025artgs,wu2026reartgs,wu2026reartgspp} require paired multi-view images of two distinct articulation states as input.
Articulation in Motion~\cite{ai2026aim} requires both multi-state observations and an interaction video showing the articulation motion.
In the rest-state setting, only a single closed-configuration observation is available, making direct application infeasible.
Rather than excluding the comparison entirely, we construct the best achievable inputs using the same video diffusion pipeline as our method, establishing an upper-bound comparison that reflects the inherent difficulty of the rest-state setting rather than any limitation of these baselines themselves.

\begin{figure}[t]
    \centering
    \includegraphics[width=\linewidth]{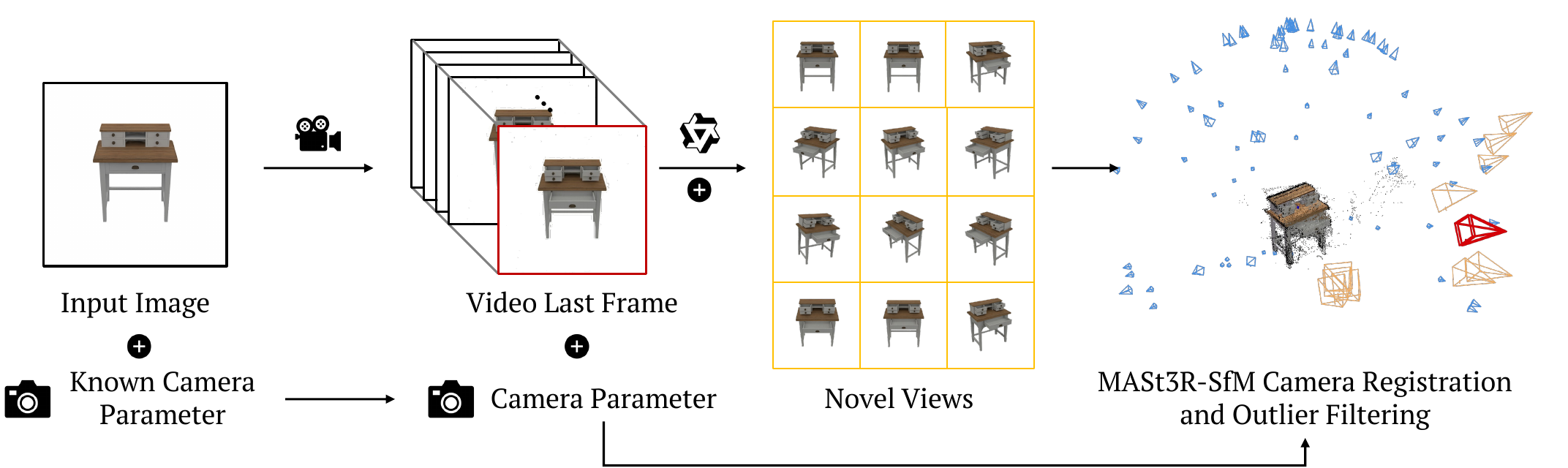}
    \caption{\textbf{Input synthesis pipeline for two-state reconstruction baselines.}
    Yellow frustums denote novel views registered via MASt3R-SfM~\cite{mast3r_eccv24,duisterhof2025mastrsfm}, and the red frustum denotes the reference view shared with the rest-state input.}
\label{fig:supp_twostate_pipeline}
\end{figure}

\subsubsection{Two-State Input Synthesis.}
We select the last physically plausible frame from the generated articulation video as the articulated-state reference.
Since our prompt enforces a strictly fixed camera throughout the video, the last frame shares the same camera parameters as the input rest-state image.
We therefore reuse the input camera as the reference pose for the articulated state.
This camera-sharing assumption also applies to our method, but multi-state methods are more sensitive to camera parameter accuracy given their reliance on precise multi-view geometry.
We therefore manually verify the absence of camera drift and hallucination artifacts for every generated video on the ACD dataset~\cite{iliash2026s2o,collins2022abo,khanna2023hssd}, retaining 46 of 69 videos before running these multi-state baselines.
Multi-view articulated-state images are then synthesized from the selected frame using Qwen-Image-Edit~\cite{wu2025qwenimagetechnicalreport,odin2025multiangleslora} with a multi-angle LoRA, which produces view-consistent images that faithfully preserve object identity and geometry across viewpoints, yielding up to 13 images per object, including the input view~(\cref{fig:supp_twostate_pipeline}).
The mask overlay is not applied to video generation in this preprocessing to maintain a consistent appearance between the rest and articulated states; this leads to higher hallucination rates compared to our method, as shown in \cref{fig:exp_ablation_video}.

\subsubsection{Camera Estimation and View Filtering.}
Camera parameters are estimated via MASt3R-SfM~\cite{mast3r_eccv24, duisterhof2025mastrsfm}, anchored to the input view as the coordinate reference.
We apply a two-pass filtering strategy to remove geometrically inconsistent views.
In the first pass, all views are registered jointly, and outlier cameras are identified using a Median Absolute Deviation~(MAD)-based robust test, where a view is rejected if its z-score exceeds $\tau = 3.0$.
Filtering is constrained to remove at most 30\% of views while retaining a minimum of 5.
In the second pass, SfM is re-run on the remaining views to improve camera accuracy.
On average, 12.54 of 13 generated views are retained per object across the 46 scenes, with outlier removal occurring in 9 scenes~(20\%).

\subsection{MultiScan Preprocessing Details}
\label{sec:supp_multiscan}
Since MultiScan~\cite{mao2022multiscan} annotations are at the scene level, we extract object-centric articulated instances as follows. 
Using the mesh-level \texttt{objectId} and \texttt{partId} labels, we crop each annotated object from the scene mesh and export its part-segmented mesh together with joint origins and axes.
We retain only openable objects whose annotations are valid.
For each instance, we then select visible RGB views by projecting its 3D points into the subsampled image frames with the provided intrinsics and poses, filter the views by visibility and scene-level occlusion, and transform the retained images and poses into the object's local frame.
Note that the two states are not always well-curated pairs of rest and articulated configurations; we retain such instances as part of the real-world capture noise rather than filtering them out.


\section{Additional Experiments and Analyses}
This section provides supplementary experimental results for \nickname.
\begin{itemize}[label=$\circ$,noitemsep,topsep=0pt,leftmargin=*]
    \item \Cref{sec:supp_convergence} analyzes the convergence behavior of the co-refinement loop.
    \item \Cref{sec:supp_vdm} analyzes the choice of video diffusion as the motion source and the robustness of joint estimation to its artifacts.
    \item \Cref{sec:supp_seg} evaluates quantitative part segmentation results against 3D segmentation baselines~\cite{liu2025partfield,ma2025find3d}.
    \item \Cref{sec:supp_failure} analyzes failure modes and directions for improvement.
\end{itemize}

\subsection{Convergence of Co-Refinement}
\label{sec:supp_convergence}

We say that the co-refinement loop \emph{converges} when the VLM's predicted part counts agree with the segmentation model's detections across all part categories, requiring no further refinement rounds.
We cap the maximum number of refinement rounds at $R_{\max}=2$~(\cref{tab:supp_impl_params}), under which $84.1\%$ of ACD scenes reach agreement.
This bound keeps inference cost low while resolving most disagreements: raising $R_{\max}$ to $20$ yields only a modest gain to $92.8\%$.

The remaining $7.2\%$ of cases do not diverge; instead, they oscillate between two stable configurations across iterations.
This behavior typically arises on objects with visual ambiguity, such as low-texture surfaces or weak part boundaries, where the VLM hierarchy and the segmentation outputs fail to settle on a single consistent decomposition.

\subsection{VDM Output as Motion Hypothesis}
\label{sec:supp_vdm}

\begin{figure}[t]
    \centering
    \includegraphics[width=0.8\linewidth]{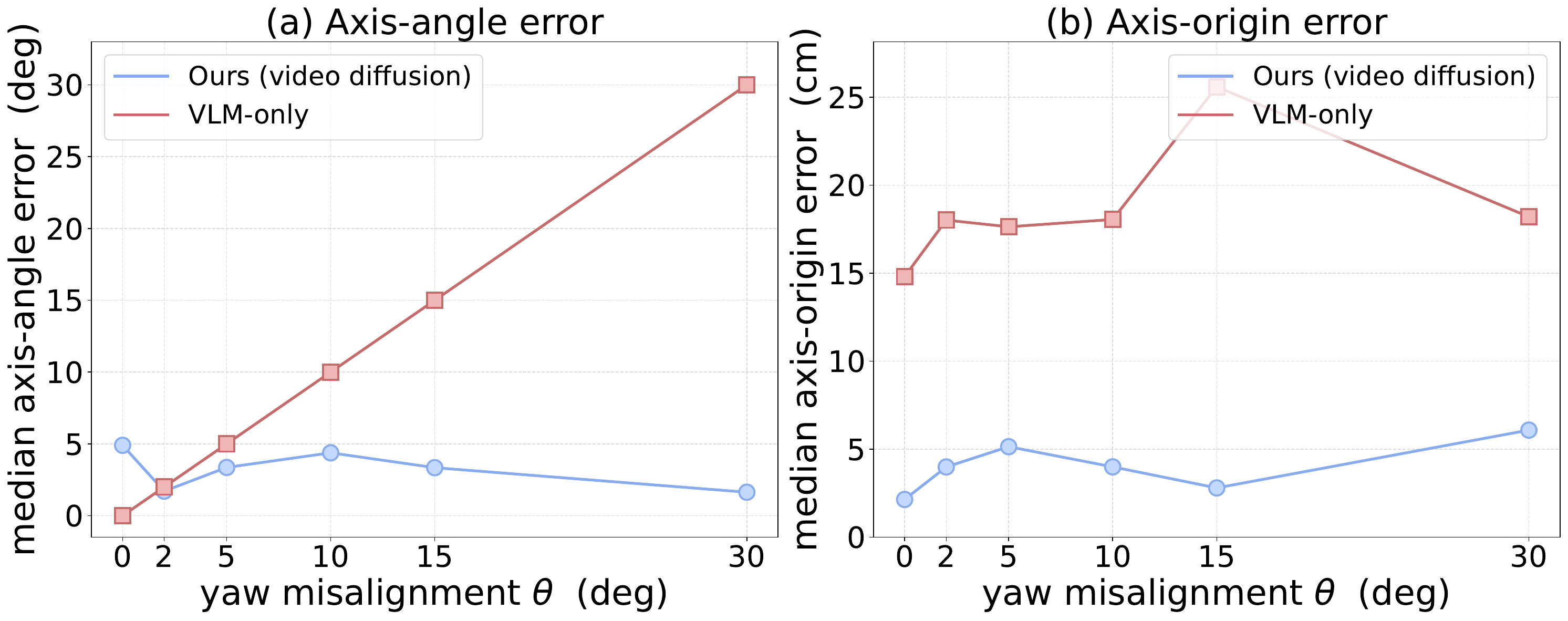}
    \caption{\textbf{Joint axis accuracy under yaw misalignment on the ACD dataset~\cite{iliash2026s2o}.}
    As the object is rotated about the vertical axis away from canonical alignment, VLM-based direct regression degrades sharply, exposing its reliance on viewpoints. Our mesh-grounded fitting remains robust.}
    \label{fig:supp_yaw}
\end{figure}

\subsubsection{Why Video Diffusion, Not a VLM.}
A natural alternative to a video diffusion model~(VDM) is to directly regress joint parameters with a vision-language model~(VLM).
However, VLMs are trained predominantly on 2D images and lack reliable 3D spatial reasoning, making direct regression of continuous joint parameters from a single rest-state image fundamentally unreliable.
Reported successes typically depend on canonical, axis-aligned viewpoints, where joint directions reduce to a small set of memorized orientations (\eg, $\pm x$, $\pm y$, $\pm z$) rather than reflecting genuine geometric inference.
We confirm this empirically by perturbing the object pose with yaw misalignment~(\cref{fig:supp_yaw}): VLM-based regression degrades sharply under even modest deviation, while our pipeline remains stable.
VDMs, by contrast, are trained on motion data and produce temporally coherent trajectories that our mesh-grounded rigid-motion fitting converts into joint parameters robust to pose noise.

\begin{figure}[t]
    \centering
    \includegraphics[width=0.45\linewidth]{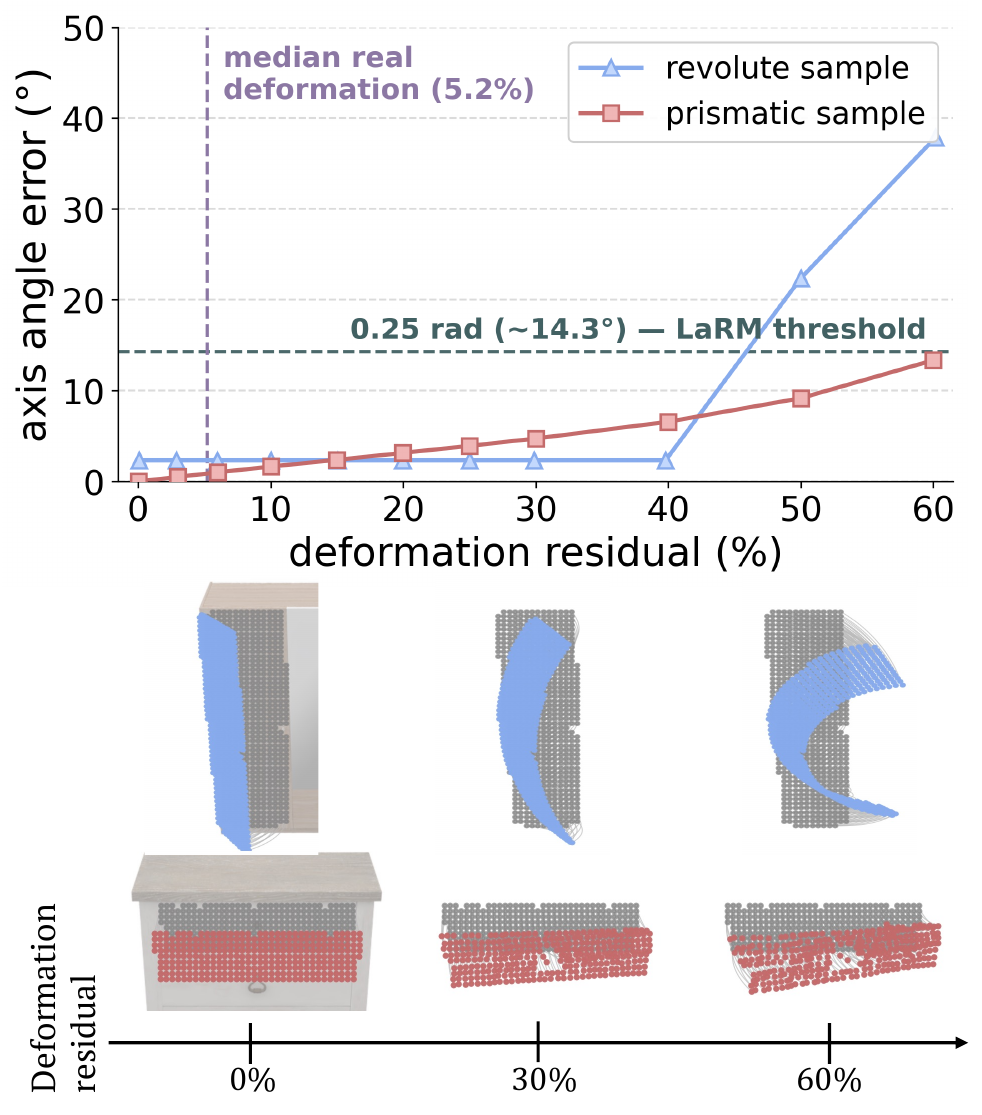}
    \caption{\textbf{Axis error under injected deformation.}
    The dashed marker indicates the empirical median deformation on the ACD dataset~\cite{iliash2026s2o}.}
    \label{fig:supp_deformation}
\end{figure}

\subsubsection{Robustness to VDM Artifacts.}
Video diffusion models produce plausible articulation trajectories from a single image, motivating their use in our pipeline.
At the same time, video generation remains an active research area, particularly in 3D-aware reasoning and physically grounded world models.
Our pipeline accordingly treats VDM output as a motion-hypothesis source and the 3D mesh as the geometric ground truth.

We use VBVR LoRA~\cite{wang2026vbvr,kijai2025vbvrlora} to suppress hallucination in generation~(\cref{fig:exp_ablation_video}), and mesh-grounded rigid-body fitting to handle unreliable trajectories and non-rigid drift.
For deformation specifically, we synthetically inject deformation into the part geometry and measure the resulting axis error~(\cref{fig:supp_deformation}).
At the empirical median deformation of $5.2\%$ on the ACD dataset~\cite{iliash2026s2o}, the axis error remains negligible, confirming that our pipeline operates effectively within the regime of realistic generation noise.

\subsection{Segmentation Quantitative Results}
\label{sec:supp_seg}

\begin{table}[t]
\centering
\caption{\textbf{Part segmentation results on ACD.}}
\label{tab:supp_seg}
\footnotesize
\setlength{\tabcolsep}{4pt}
\resizebox{0.45\linewidth}{!}{
\begin{tabular}{l ccc}
\toprule
 & Find3D~\cite{ma2025find3d} & PartField~\cite{liu2025partfield}  & \textbf{Ours} \\
\midrule
IoU & 0.20 & 0.21 & \textbf{0.88} \\
mAP & 0.70 & 0.46 & \textbf{0.89} \\
\bottomrule
\end{tabular}
}
\end{table}

We evaluate part segmentation quality on the ACD dataset~\cite{iliash2026s2o,collins2022abo,khanna2023hssd} using IoU and mAP, comparing against 3D segmentation methods PartField~\cite{liu2025partfield} and Find3D~\cite{ma2025find3d}.
\nickname substantially outperforms both baselines across all metrics, demonstrating the benefit of grounding segmentation in the reconstructed mesh geometry combined with the VLM-predicted part hierarchy.
Find3D achieves relatively high mAP despite low IoU, as text-prompted parts are often under-segmented, leaving a large unsegmented base region that dominates the evaluation and inflates mAP.

\subsection{Failure Case Analysis}
\label{sec:supp_failure}
Although our co-refinement pipeline with mask overlay suppresses hallucination artifacts, 2 out of 69 videos (2.9\%) still exhibit corrupted geometry that prevents meaningful track extraction.
Degraded joint accuracy primarily occurs on objects with subtle or short-range articulation, where track displacement is too small to reliably estimate motion parameters.
We see two directions for improvement.
First, increasing the number of generated videos enlarges the hypothesis pool, improving the chance of capturing reliable motion for difficult cases.
Second, since our pipeline leverages 3D geometric cues such as collision and spatial constraints for joint refinement, the quality of the underlying mesh directly affects estimation accuracy. Cleaner part boundaries from improved segmentation, potentially assisted by generative remeshing models, would reduce ambiguity in the geometric reasoning stage.

\section{Applications}
This section demonstrates \nickname beyond the standard evaluation setting.
\begin{itemize}[label=$\circ$,noitemsep,topsep=0pt,leftmargin=*]
    \item \Cref{sec:supp_single_state} shows that \nickname generalizes to articulated-state inputs without any modification.
    \item \Cref{sec:supp_realworld} validates \nickname in real-world captures across diverse input modalities and articulation states.
    \item \Cref{sec:supp_generalization} extends \nickname to broader object categories through a meta-prompt mechanism.
    \item \Cref{sec:supp_downstream} demonstrates downstream applications of \nickname.
\end{itemize}

\subsection{Single Articulated-State Input}
\label{sec:supp_single_state}

\begin{figure}[t]
    \centering
    \includegraphics[width=0.55\linewidth]{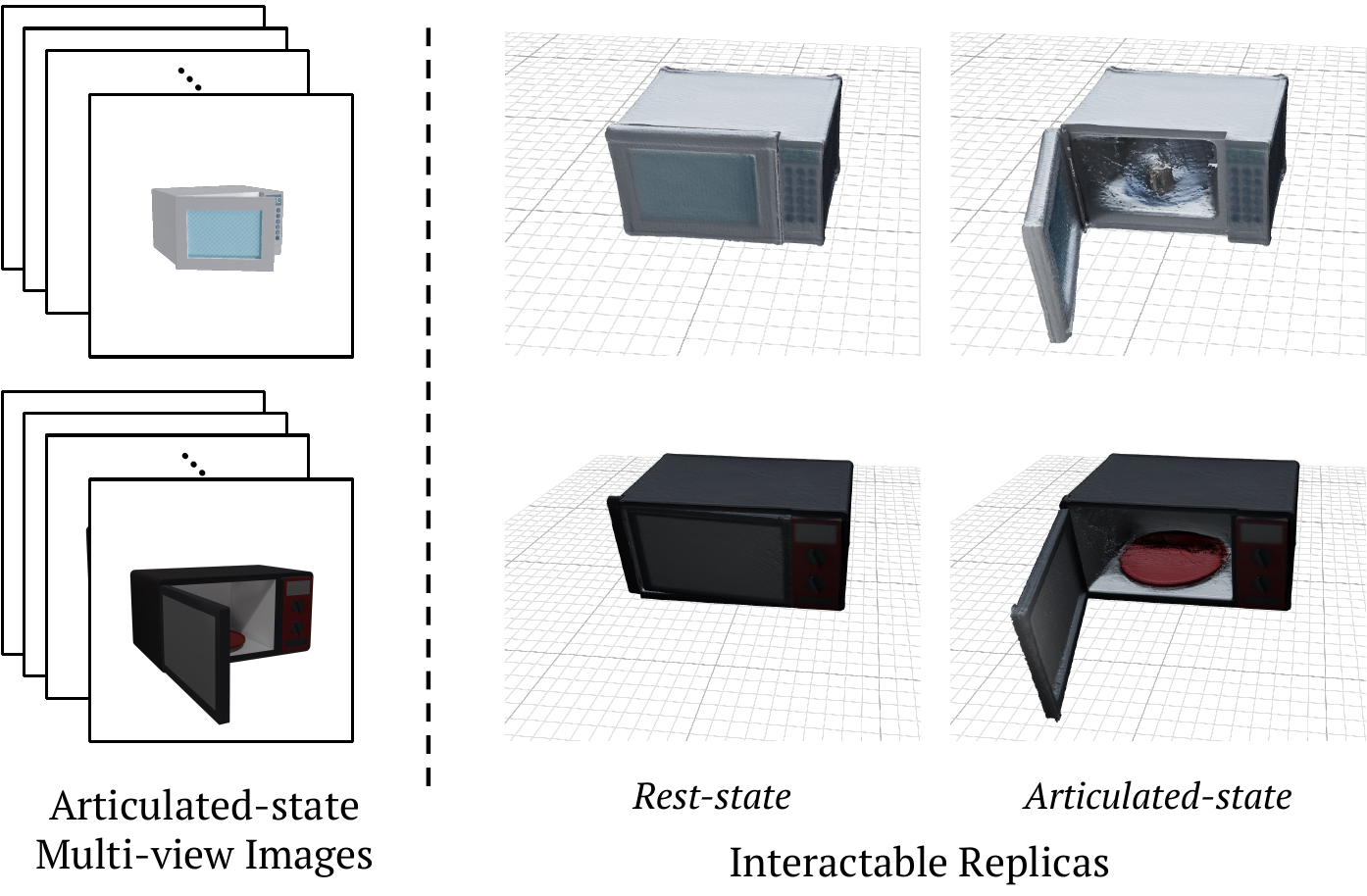}
    \caption{\textbf{Results on articulated-state inputs from the PartNet-Mobility \textit{microwave} category.}}
    \label{fig:supp_single}
\end{figure}

Although \nickname is designed for rest-state observation, the same pipeline generalizes to objects observed in a partially or fully open configuration, as shown in \cref{fig:supp_single}.
This is because rest-state reconstruction is strictly harder: the closed configuration provides no direct motion cues, requiring the full pipeline, including motion hypothesis generation and mesh-grounded refinement, to recover joint parameters from appearance alone.
An articulated-state observation, by contrast, already encodes part displacement in the input geometry, so the same pipeline succeeds with less reliance on the hypothesis generator.
As a result, any input that \nickname handles in the rest-state setting is also handled in the articulated-state setting, but not vice versa.
We demonstrate this on \textit{microwave} instances from the PartNet-Mobility dataset~\cite{xiang2020sapien,mo2019partnet,chang2015shapenet}, omitting mesh completion as the articulated state directly exposes interior geometry.

\begin{figure}[t]
    \centering
    \includegraphics[trim={8mm 0mm 0mm 0mm}, clip, width=0.8\linewidth]{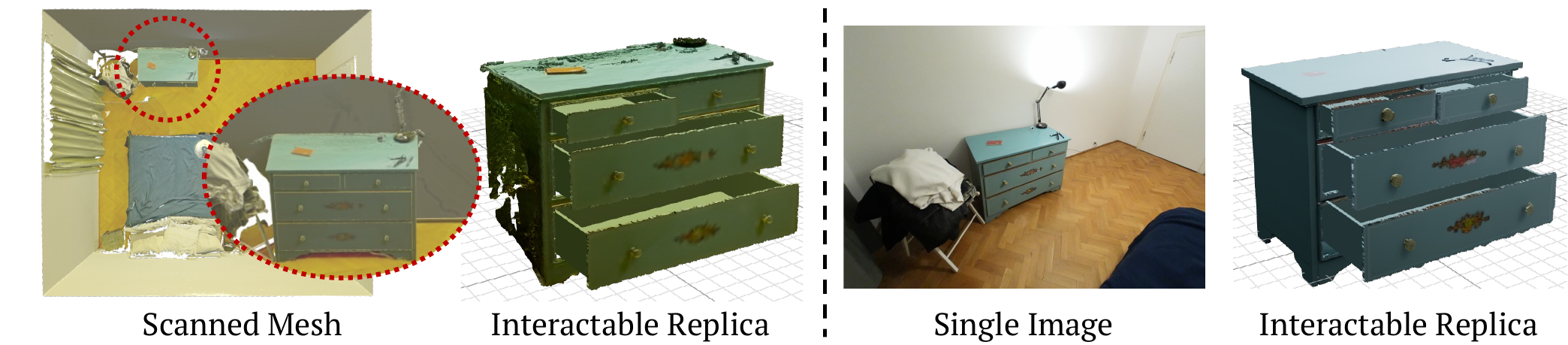}
    \caption{\textbf{Real-world results on a ScanNet${++}$ scene~\cite{yeshwanthliu2023scannetpp}.} \textbf{Left:} a raw scanned mesh. \textbf{Right:} a single captured image reconstructed via SAM3D~\cite{sam3dteam2025sam3d3dfyimages}.}
    \label{fig:supp_scannet}
\end{figure}

\begin{figure}[t]
    \centering
    \includegraphics[width=0.8\linewidth]{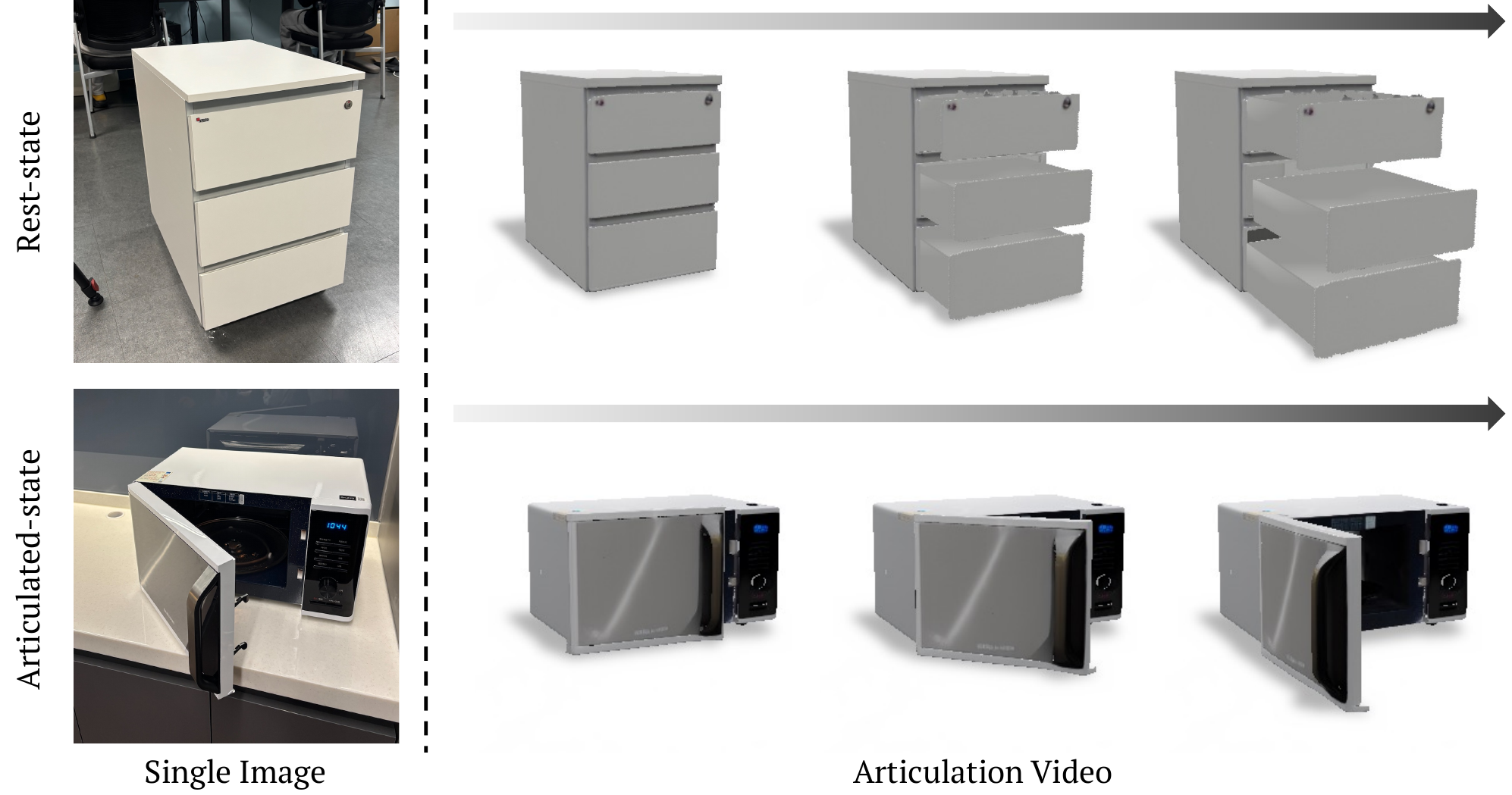}
    \caption{\textbf{Real-world single-image results.} Given a single casually captured image, \nickname reconstructs the object via SAM3D~\cite{sam3dteam2025sam3d3dfyimages} and estimates articulation. \textbf{Top:} A rest-state cabinet. \textbf{Bottom:} A microwave captured in a partially open configuration.}
    \label{fig:supp_realworld}
\end{figure}

\subsection{Real-World Demonstrations}
\label{sec:supp_realworld}

We demonstrate \nickname on a diverse set of real-world captures spanning multiple input modalities.
\Cref{fig:supp_scannet} presents results on the same ScanNet${++}$ scene~\cite{yeshwanthliu2023scannetpp} under two modalities: a raw scanned mesh and a single captured image reconstructed via SAM3D~\cite{sam3dteam2025sam3d3dfyimages}, showing consistent articulation estimates across reconstruction pipelines.
\Cref{fig:supp_realworld} shows results on casually captured in-the-wild images, covering both rest-state and partially open inputs.
In all cases, \nickname exports simulation-ready URDF assets suitable for embodied AI and robotics applications.

\begin{figure}[t]
    \centering
    \includegraphics[width=0.5\linewidth]{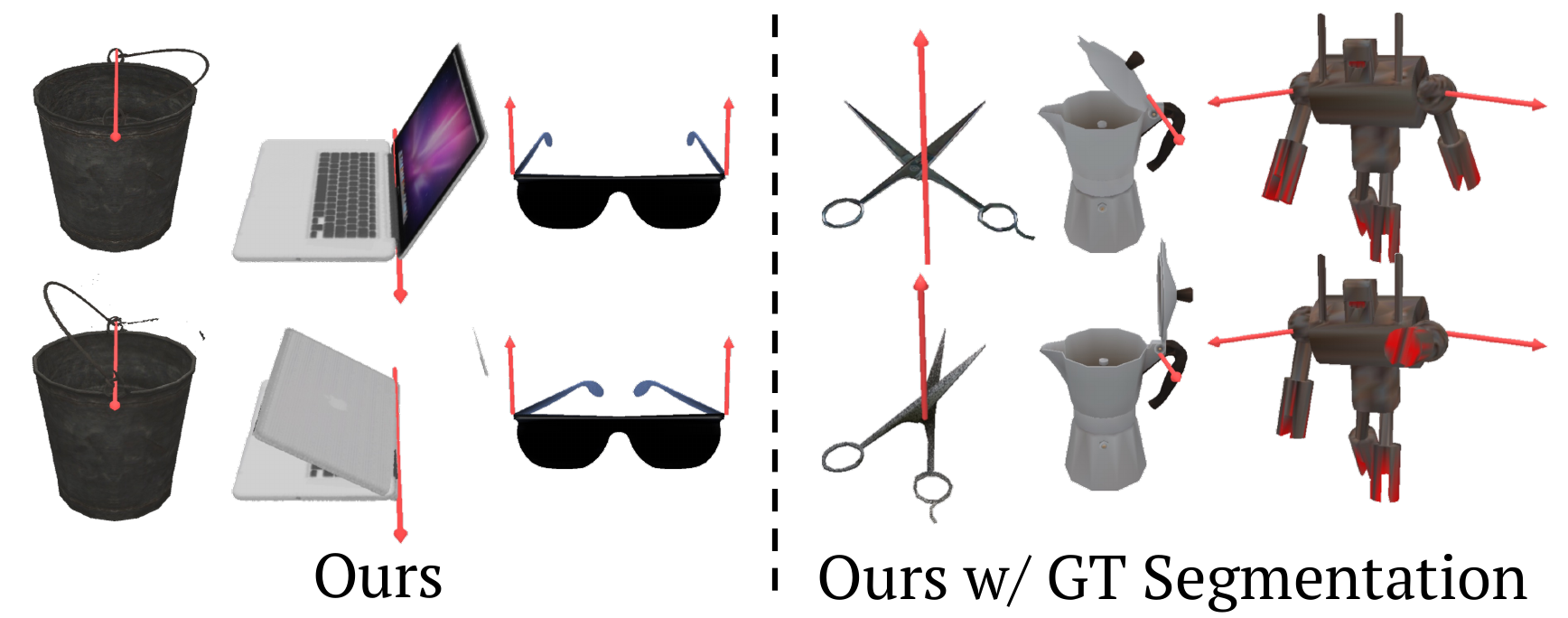}
    \caption{\textbf{Generalization to broader object categories via meta-prompting.}}
    \label{fig:supp_general_obj}
\end{figure}

\begin{figure}[t]
    \centering
    \includegraphics[width=0.8\linewidth]{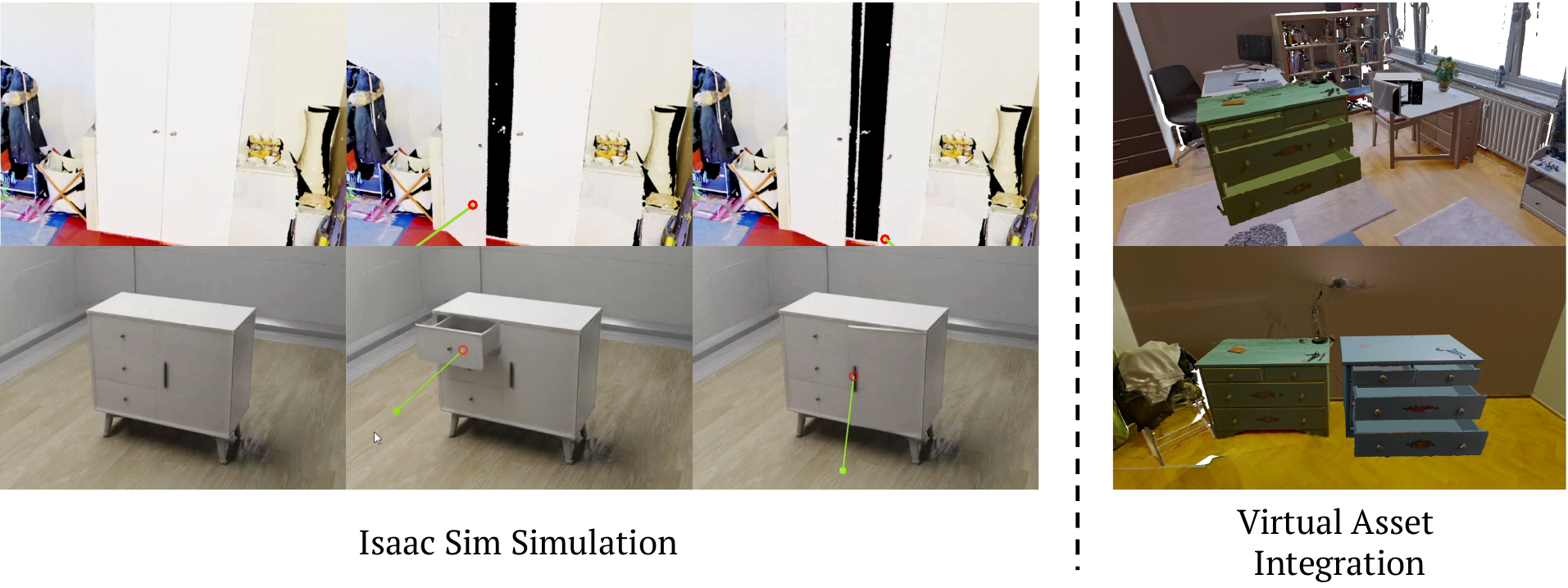}
    \caption{\textbf{Downstream applications of \nickname.} \textbf{Left:} Physics-based manipulation in Isaac Sim~\cite{NVIDIA_Isaac_Sim}, showing a reconstructed cabinet with operable drawers. \textbf{Right:} Virtual asset integration, placing a reconstructed interactable replica into a real-world scene.}
    \label{fig:supp_app}
\end{figure}

\subsection{Generalization to General Object Categories}
\label{sec:supp_generalization}

While our main evaluation targets openable furniture with prismatic and revolute joints, our framework generalizes to broader categories through a \emph{meta-prompt} mechanism.
Articulated objects span a wide range of part vocabularies (\eg, a laptop's display lid, a robot's arm) that no single fixed prompt covers well.
Our pipeline therefore issues two VLM calls: the first sends a category-agnostic system prompt and asks the VLM to rewrite it for the input object, adapting the part vocabulary, the per-object definition of ``base,'' and the example response.
The second call applies the rewritten prompt to the same image and produces the part-connectivity tree, which feeds the unchanged mask, segmentation, video, and joint-fitting modules.
\Cref{fig:supp_general_obj} shows representative results on PartNet-Mobility~\cite{xiang2020sapien,mo2019partnet,chang2015shapenet} categories~(eyeglasses, laptops, scissors, buckets) and Objaverse~\cite{objaverse} assets~(moka pot, mechanical robots).
Extending to non-planar joints and more general articulation models remains an open direction.

\subsection{Downstream Applications}
\label{sec:supp_downstream}
The simulation-ready URDF assets produced by \nickname can be directly deployed in downstream applications including physics simulation and virtual asset integration, as shown in \cref{fig:supp_app}.

%% file: tables/x_supp_param.tex
\begin{table}[h]
\centering
\setlength{\tabcolsep}{4pt}
\caption{\textbf{Implementation parameters of \nickname.}
Camera parameters apply only when the input is a standalone mesh;
for NVS-based inputs, intrinsics are inherited from training cameras and only extrinsics follow the placement rules in \cref{sec:supp_view}.
}
\resizebox{0.8\columnwidth}{!}{
\begin{tabular}{clc}
\toprule\midrule
\textbf{Param.} & \textbf{Description} & \textbf{Value} \\
\midrule
\multicolumn{3}{l}{\textit{View Selection}} \\
$N'$               & Total re-rendered views                        & 12   \\
$R_\text{max}$     & Max co-refinement rounds                       & 2    \\
\midrule
\multicolumn{3}{l}{\textit{Camera (mesh input only)}} \\
$-$                & Rendering resolution                           & $800{\times}800$ \\
$-$                & Field of view                                  & $35.49°$ \\
$-$                & Camera radius                                  & 4.5  \\
$-$                & Azimuth range from frontal ($-y$)              & $\pm 30°$ \\
$-$                & Elevation range                                & $15°$--$45°$ \\
\midrule
\multicolumn{3}{l}{\textit{Articulation Estimation}} \\
$R$                & Number of video generation seeds               & 2    \\
$\tau_\text{vis}$  & Visibility threshold for track truncation      & 0.5  \\
\midrule\bottomrule
\end{tabular}
}
\label{tab:supp_impl_params}
\end{table}

%% file: algorithms/supp_vlm_prompt_box.tex
\begin{tcolorbox}[
    colback=gray!8,
    colframe=gray!50,
    boxrule=0.4pt,
    arc=2pt,
    left=6pt, right=6pt, top=4pt, bottom=4pt,
    fontupper=\scriptsize 
]
You are an expert in the recognition of articulated parts of an object in an image.
You will be provided with an image of an articulated object.

\medskip
\noindent You should follow the following steps to achieve the task:

\noindent(1) Recognize all the articulated parts of the object from the set
\{base, door, knob, handle, drawer, tray\}.
There must always be exactly one base.
Trays can only exist in microwaves.
Each handle must be attached to either a door or a drawer.
Each door can have at most two handles.

\medskip
\noindent To distinguish between a door and a drawer, reason carefully about handle placement, panel proportion, and likely motion affordance: doors provide access to internal compartments and may be hinged or sliding; drawer fronts are box-like units that slide outward.
Do not rely on a single rigid rule; use overall geometry, proportion, and functional affordance.

\medskip
\noindent(2) Describe how the parts are connected and organize them into a part connectivity graph.
The base must always be the root.
All other parts must be attached hierarchically under the base or another articulated part.

\medskip
\noindent{Output format:} Include a valid JSON object within ```json ...\ ''' fences with exactly one root key ``base''.
Output only the textual description followed by the JSON block.
\end{tcolorbox}

%% file: algorithms/supp_vdm_prompt.tex
\begin{tcolorbox}[
    colback=gray!8,
    colframe=gray!50,
    boxrule=0.4pt,
    arc=2pt,
    left=6pt, right=6pt, top=4pt, bottom=4pt,
    fontupper=\scriptsize 
]
You are an expert in articulated object reasoning and motion-aware prompt generation.
Given a single image of an object, write ONE WAN2.2 video-generation prompt that forces ALL articulated parts to move simultaneously while the camera remains completely fixed.

\medskip
\noindent\textit{Requirements:}

\noindent(1) The camera is strictly locked for the entire clip: fixed intrinsics and extrinsics, no pan/tilt/roll/zoom/dolly, no stabilization drift, no parallax, no reframing.

\noindent(2) All articulated parts move at the same time: drawers slide outward together, doors/panels swing open together, handles move only with their attached parts. Motions are synchronized, smooth, mechanically plausible, moderate range, gravity-consistent, and collision-free.

\noindent(3) Preserve exact object identity and geometry; do not add, remove, or duplicate parts. Keep textures, patterns, background, and lighting consistent across frames.

\noindent(4) Photorealistic, soft studio lighting, stable geometry, cinematic clarity.

\medskip
\noindent Do NOT mention segmentation masks, colors, or annotations.\\
Output only the final WAN2.2 prompt.
\end{tcolorbox}

%% file: tables/x_supp_runtime.tex
\setlength{\intextsep}{-15pt}
\begin{wrapfigure}{r}{0.39\linewidth}
\centering
\captionsetup{type=table}
\captionof{table}{\textbf{Per-stage runtime on ACD dataset.}}
\label{tab:supp_runtime}
\footnotesize
\setlength{\tabcolsep}{3.7pt}
\resizebox{0.8\linewidth}{!}{
\begin{tabular}{lc}
\toprule\midrule
\textbf{Stage} & \textbf{Mean (s)} \\
\midrule
Co-refinement  & 30  \\
Segmentation   & 3   \\
Prompt Gen.    & 18  \\
Video Gen.     & 172 \\
Tracking       & 6   \\
Joint Fitting  & 7   \\
Mesh Completion& 40  \\
\midrule
\textbf{Total} & \textbf{$\sim$277} \\
\midrule\bottomrule
\end{tabular}}

\includegraphics[trim={0 0 0 -3em}, clip, width=0.8\linewidth]{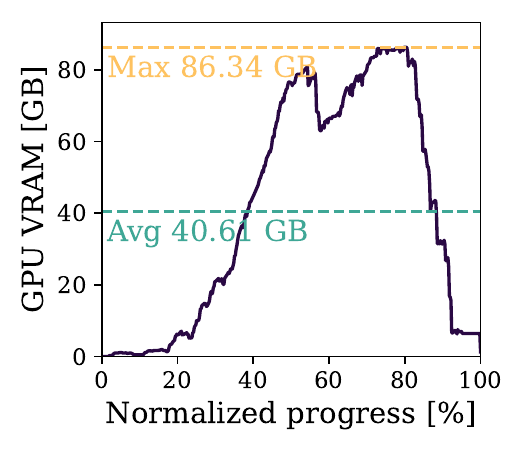}
\captionof{figure}{\textbf{Per-stage VRAM usage on ACD dataset.}}
\label{fig:supp_vram}

\end{wrapfigure}